%% file: main.tex
\documentclass[letterpaper]{article} 
\usepackage{aaai25}  
\usepackage{times}  
\usepackage{helvet}  
\usepackage{courier}  
\usepackage[hyphens]{url}  
\usepackage{graphicx} 
\usepackage{natbib}  
\usepackage{caption} 
\usepackage{algorithm}
\usepackage{algorithmic}
\usepackage{subfigure}
\usepackage{multirow}
\newcommand{\ul}[1]{\underline{#1}}
\newcommand{\bo}[1]{\boldsymbol{#1}}

\usepackage{amssymb}
\usepackage{amsmath}
\usepackage{booktabs}
\usepackage{cleveref}
\usepackage{makecell}
\usepackage{newfloat}
\usepackage{listings}
\DeclareCaptionStyle{ruled}{labelfont=normalfont,labelsep=colon,strut=off} 
\floatstyle{ruled}
\newfloat{listing}{tb}{lst}{}
\floatname{listing}{Listing}
\title{TimeDP: Learning to Generate Multi-Domain Time Series with Domain Prompts}

\author{
  Yu-Hao Huang\textsuperscript{\rm 1}, Chang Xu\textsuperscript{\rm 2}\thanks{Corresponding author.}, Yueying Wu\textsuperscript{\rm 3}, Wu-Jun Li\textsuperscript{\rm 1}, Jiang Bian\textsuperscript{\rm 2}}
  
\affiliations{
  \textsuperscript{\rm 1}National Key Laboratory for Novel Software Technology,
	School of Computer Science, Nanjing University\\ \ \ \textsuperscript{\rm 2}Microsoft Research Asia \\
  \textsuperscript{\rm 3}Peking University \\
  huangyh@smail.nju.edu.cn, \{chanx, jiang.bian\}@microsoft.com \\
  wuyueying@stu.pku.edu.cn, liwujun@nju.edu.cn
}

\input{math_def.tex}

\begin{document}
\maketitle

\begin{abstract}
  \input{src_chapters/0_abs.tex}
\end{abstract}

\begin{links}
    \link{Code}{https://github.com/YukhoY/TimeDP}
\end{links}

\input{src_chapters/1_intro_new.tex}

\input{src_chapters/2_related.tex}
\input{src_chapters/3_method.tex}
\input{src_chapters/4_exps.tex}
\input{src_chapters/5_conclusion.tex}

\clearpage
\section{Acknowledgment}
This work is supported by  NSFC
 Project~(No.62192783, No.12326615) 
\bibliography{main.bib}

\include{src_chapters/appendix.tex}

\end{document}

%% file: math_def.tex
\def\bcP{{\boldsymbol{\mathcal{P}}}}
\def\bsm{{\boldsymbol{m}}}
\def\bsx{{\boldsymbol{x}}}
\def\bsz{{\boldsymbol{z}}}
\def\bsK{{\boldsymbol{K}}}
\def\bsQ{{\boldsymbol{Q}}}
\def\bsV{{\boldsymbol{V}}}
\def\bsW{{\boldsymbol{W}}}
\def\bsP{{\boldsymbol{P}}}
\def\bsp{{\boldsymbol{p}}}

\def\bep{{\boldsymbol{\epsilon}}}

%% file: src_chapters/0_abs.tex
Time series generation models are crucial for applications like data augmentation and privacy preservation. 
Most existing time series generation models are typically designed to generate data from one specified domain. 
While leveraging data from other domain for better generalization is proved to work in other application areas, this approach remains challenging for time series modeling due to the large divergence in patterns among different real world time series categories. 
In this paper, we propose a multi-domain \ul{time} series \ul{d}iffusion model with domain \ul{p}rompts, named \textbf{TimeDP}. 
In TimeDP, we utilize a time series semantic prototype module which defines time series prototypes to represent time series basis, each prototype vector serving as ``word'' representing some elementary time series feature. 
A prototype assignment module is applied to extract the extract domain specific prototype weights, for learning domain prompts as generation condition.
During sampling, we extract ``domain prompt" with few-shot samples from the target domain and use the domain prompts as condition to generate time series samples.
Experiments demonstrate that our method outperforms baselines to provide the state-of-the-art in-domain generation quality and strong unseen domain generation capability.

%% file: src_chapters/1_intro_new.tex
\section{Introduction}

In the landscape of large models and advanced machine learning techniques, time series foundation models~\cite{timesfm, units} have garnered increasing attention. These models, typically trained on extensive datasets spanning various domains~\cite{moirai}, have predominantly emphasized forecasting tasks rather than the generation of new data. However, the accurate and meaningful generation of time series is critical for applications such as medical record synthetic~\cite{li23medi} and financial scenario simulations~\cite{coletta21, diga}, as well as for augmenting datasets where historical records are limited or incomplete~\cite{kollovieh2023predict}. 

Although some research has been conducted on time series generation, most efforts have been confined to the development of generation model for single-domain data. In contrast, cross-domain time series generation presents a significantly more complex challenge, as it requires the creation of new data across various domains without relying on existing historical records. This stands out as a gap, underscoring a substantial opportunity for further advancements in multi-domain time series generation.

One straightforward approach to multi-domain time series generation involves the use of predefined domain labels during the training process~\cite{timevq}. This method relies on the availability of domain labels to formulate the conditional generation process. However, this approach may struggle generalizing to large number of domains or unseen domains. Moreover, the challenge intensifies when domain labels are not explicitly available.

An alternative approach frames cross-domain time series generation as a conditional generation task by describing the domain using natural language~\cite{timellm, tsllmsurvey}. 
However, the use of natural language descriptions introduces significant challenges. Domain-specific nuances are often difficult to articulate precisely, leading to noisy, incomplete, or ambiguous prompts. Moreover, for entirely new or evolving domains, crafting these domain descriptors can be impractical. This has underscored a critical need for a more systematic and robust way to represent and utilize domain-specific information in time series generation.

To address these challenges, we propose a label-free, text-free method that learns time series prototypes as basic elements to construct domain prompts for generating time series with a diffusion model, named \textbf{TimeDP}. Through training, the prototypes learn to represent time series basis, serving as ``word" with time series semantics. 
A prototype assignment module is applied for each training samples to construct the specific ``prompt" for generating this sample.
During sampling, we extract ``prompt" with few-shot samples from the target domain to construct the population of domain prompts and use the domain prompts as condition to generate time series samples. 

To summarize, the main contributions of this paper are listed as follows:
\begin{itemize}
    \item We propose \textbf{TimeDP}, a multi-domain time series generation model by learning a set of time series prototypes and prototype assignment module to construct domain prompts, where the domain prompts serve as condition for a time series diffusion model. 
    \item We are the first to propose a multi-domain time series generation model using label-free, text-free conditioning mechanism.
    \item Experiments demonstrate that our method outperforms baselines with the state-of-the-art in-domain generation quality, and strong unseen domain generation capability.
\end{itemize}

%% file: src_chapters/2_related.tex
\section{Related Work and Backgrounds}

\subsection{Time Series Generation}
Existing time series generation models has based on various foundational type of generative models. 
GAN-based methods has been introduced to encourage the network to consider temporal dynamic by jointly optimize both supervised and adversarial objectives for a learned embedding space~\citep{timegan}. 
VAE-based methods have designed specific decoder structure for temporal data considering trend and seasonal decomposition~\citep{timevae}, and first introduces vector quantization technique together with bidirectional transformers to better capture temporal consistency~\citep{timevq}. 
Another category is considered as mixed-type methods, combining GANs, flows and ODEs~\citep{gtgan}. 
Different from these methods, we utilize denoising diffusion probabilistic models~(DDPM) as our generation backbone.

Existing diffusion-based time series generation methods leverage both unconditional and conditional diffusion models for generating time series data with various denoising network backbones~\citep{tsdiff,tsdsur}. 
Researchers have also considered combining diffusion models with the constrained generation problem~\citep{diffcog} and the extraction of time series intrinsic such as seasonal-trend decomposition techniques~\citep{diffts}. 
Compared with these single-domain methods, we first propose to utilize label-free, text-free domain prompts as condition for generating time series.

\subsection{Cross-Domain Time Series Model}
There have been several recent work consider utilizing multiple-domain time series data for training time series foundation models. These works can be divided into two branches. The first branch builds two-stage models. \citet{xit} pretrains a representation learning model on 75 datasets for the first stage and finetuning to task specific models at the second stage. \citet{units} conducts masked reconstruction pretraining on 38 multi-domain datasets for the first stage and a multi-task supervised learning for downstream tasks. The second branch pretrains end-to-end transformer models with patch tokenizers for time series forecasting. \citet{moirai} pretrains on a dataset with over 2B observations and \citet{timesfm} pretrains on a dataset with 100B time-points, both using patching and instance normalizing to unify across different data scale, frequencies and lengths. These methods employ instance normalization to generate forecasts based on historical data without explicitly addressing domain differences. Compared with these approaches, we propose to use time series prototypes, constructing domain prompts to explicitly distinguish domains as well as bridge them.

\subsection{Denoising Diffusion Probabilistic Models~(DDPMs)}
A diffusion probabilistic model~\citep{SD15} learns to reverse the transitions of a Markov chain which is known as the diffusion process that gradually adds noise to data, ultimately destroying the signal. 

Let $\mathbf{x}_0 \in \mathbb{R}^d \sim q(\mathbf{x}_0)$ be real data of dimension $d$ from space $\mathcal{X}$. The diffusion process generates $\mathbf{x}_1, ..., \mathbf{x}_N$ from the same space with the same shape as $\mathbf{x}_0$, using a Markov chain that adds Gaussian noise over $N$ time steps: $q(\mathbf{x}_1,...,\mathbf{x}_N|\mathbf{x}_0):=\prod_{n=1}^N{q(\mathbf{x}_n|\mathbf{x}_{n-1})}$. The transition kernel is commonly defined as:
\begin{equation}\label{eq:fw}
    q(\mathbf{x}_n|\mathbf{x}_{n-1}):=\mathcal{N}(\mathbf{x}_n;\sqrt{1-\beta_n}\mathbf{x}_n,\beta_n\mathbf{I}),
\end{equation}
where $\{ \beta_n \in (0,1)\}_{n=1,...,N}$ defines the variance schedule. Note that $\mathbf{x}_n$ at any arbitrary time step $n$ can be derived in a closed form $q(\mathbf{x}_n|\mathbf{x}_0)=\mathcal{N}(\mathbf{x}_n;\sqrt{\bar{\alpha}_n}\mathbf{x}_0,(1-\bar\alpha_n)\mathbf{I})$, where $\alpha_n:=1-\beta_n$ and $\bar\alpha_n:=\prod^n_{s=1}\alpha_s$. For the reverse process, the diffusion model, parameterized by $\theta$, yields:
\begin{equation}
p_\theta(\mathbf{x}_0,\mathbf{x}_1,...,\mathbf{x}_N):=p(\mathbf{x}_N)\prod_{n=1}^N{p_\theta(\mathbf{x_{n-1}|\mathbf{x}_n})},
\end{equation}
where 
$p_\theta(\mathbf{x}_{n-1}|\mathbf{x}_n):=\mathcal{N}(\mathbf{x}_{n-1};\boldsymbol{\mu}_\theta(\mathbf{x}_n,n),\boldsymbol{\Sigma}_\theta(\mathbf{x}_n,n))$ 
and the transitions start at $p(\mathbf{x}_N)=\mathcal{N}(\mathbf{x}_n;\mathbf{0},\mathbf{I})$. 
The optimization objective is derived into the maximizing of an approximation of the evidence lower bound~(ELBO) of the log-likelihood $\log p_\theta(\mathbf{x}_0)$.
With a widely adopted parameterization:
\begin{equation}
    \boldsymbol{\mu}_\theta(\mathbf{x}_n,n)=\frac{1}{\sqrt{\alpha_n}}(\mathbf{x}_n-\frac{\beta_n}{\sqrt{1-\bar\alpha_n}}\boldsymbol{\epsilon}_\theta(\mathbf{x}_n,n)),
\end{equation}
the training is performed to predict the noise term added in the forward process which simplifies the objective to:
\begin{equation}\label{uncond_diff}
L_{simple}:=\mathbb{E}_{\mathbf{x}_0,\boldsymbol{\epsilon},n}[\|\boldsymbol{\epsilon}-\boldsymbol{\epsilon_\theta}(\sqrt{\bar\alpha_n}\mathbf{x}_0+\sqrt{1-\bar\alpha_n}\boldsymbol{\epsilon},n)\|^2]. 
\end{equation}
On sampling, $\mathbf{x}_{n-1}=\frac{1}{\sqrt{\alpha_n}}(\mathbf{x}_n-\frac{1-\alpha_n}{\sqrt{1-\bar\alpha_n}}\boldsymbol{\epsilon}_\theta(\mathbf{x}_n,n))+\sigma_n\mathbf{z}$, where~$\sigma_n = \sqrt{\beta_n}$ and $\mathbf{z} \sim \mathcal{N}(\mathbf{0},\mathbf{I})$ \cite{ho20}.

Typical time series diffusion models for forecasting task~\cite{timegrad,timediff,mgtsd} encodes history context into $c$ as condition and make use of the conditional form of DDPMs~\cite{ho21} for generating future time series:
\begin{equation}\label{eq:cond}
p_\theta(\mathbf{x}_0,\mathbf{x}_1,...,\mathbf{x}_N|c):=p(\mathbf{x}_N)\prod_{n=1}^N{p_\theta(\mathbf{x_{n-1}|\mathbf{x}_n},c)},
\end{equation}
In the problem setting of time series generation, the generation process does not rely on time series history. We explore the use of term $c$ to provide domain semantics for time series generation model in this work.

\input{src_figtabs/fig_method.tex}

\subsection{Problem Formulation}
Let $D^T_i=\{\boldsymbol x \in \mathbb{R}^{T}\}^{N_i}, \boldsymbol{x} = (x_1,x_2,...,x_T)$ denote a time series dataset domain $i$ with $N_i$ time series samples, where each sample contains $T$ sequential values. A straightforward single-domain time series generation model fits the joint distribution over time steps $p(x_1, x_2, ..., x_T)$ of each dataset with a separate model parameterized by $\theta_i$, namely $p_{\theta_i}(x_1,x_2,...,x_T)$ for all $\boldsymbol{x}$ in $D^T_i$. 

In this paper, we explore a domain-unified setting where the mixture of $M$ domain datasets with sequence length $T$ is denoted by $D^T=\bigcup^M_{i=1}D_i$ and we aim to build one model for the mixed dataset parameterized by $\theta$, namely $p_\theta(x_1,x_2,...,x_T|i)$ for all $\boldsymbol{x}$ in $D^T$.

Adhering to the channel-independent setting~\citep{patchtst} that is widely accepted by recent researches, we formulate the problem studied in this paper in a uni-variate time series generation manner to handle the heterogeneity of time series in terms of dimension~\citep{moirai}. 

%% file: src_figtabs/fig_method.tex
\begin{figure*}[!t]
  \centering
  \includegraphics[width=0.95\textwidth]{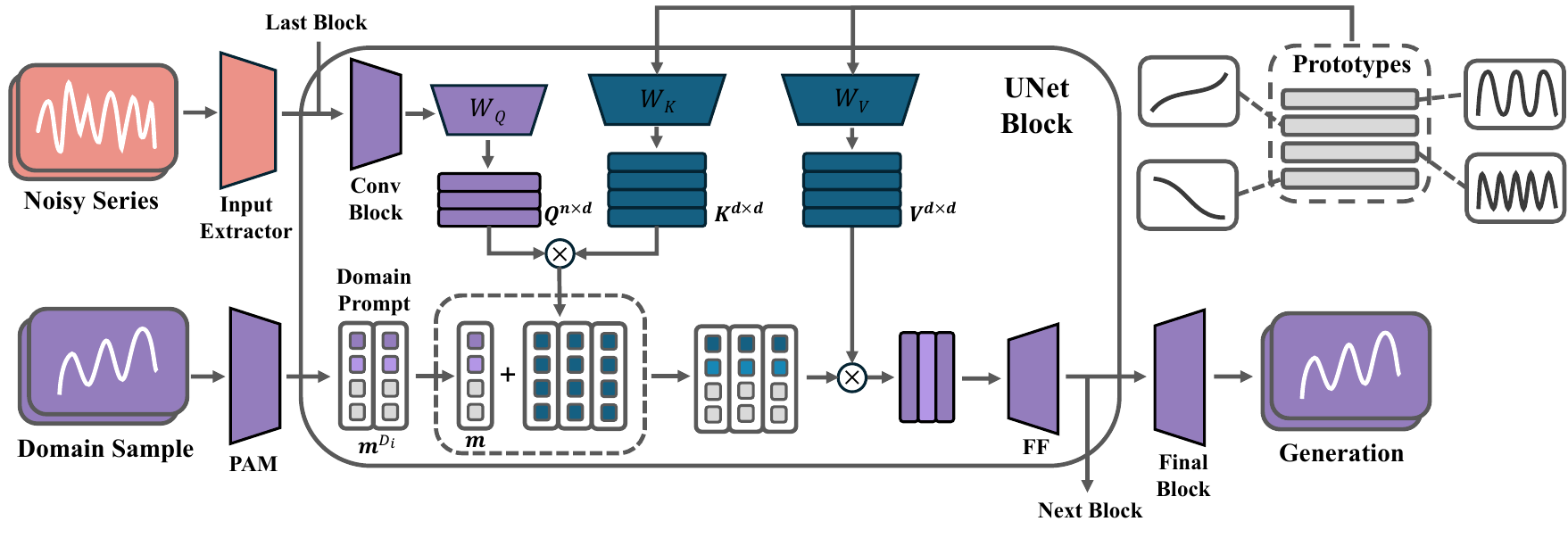}
  \caption{Overview of TimeDP model.}
  \label{fig:overview}
\end{figure*}

%% file: src_chapters/3_method.tex
\section{Methodology}
With sequences from all data domains mixed together during training, all time series features within latent representation are entangled without a explicit way for distinguishing specific time series data domain. 
Although utilizing domain labels as class labels for training the time series generation model can provide instruction for identifying specific domain, this approach implies an assumption that all domains are independent from each other, neglecting the different similarity levels among domain pairs. Therefore, it is challenging to equip the model with the ability of generating time series in selected domain while considering the inter-relationship among domains.
To overcome this challenge, the key is to build a triggering mechanism for cross-domain time series model that can control the model for generating time series data from specific domain. 
Motivated by the recent advancements in controllable content generation with prompting technique, we propose to construct domain prompts for controlling a cross-domain model.

In the rest of this section, we first describe the model architecture design. Then, we describe the optimization objective and training algorithm of the proposed model. Finally, we discuss the procedure for in-domain sampling and unseen domain generation using the proposed model.

\subsection{Domain Prompts}
\label{sec:semdec}
Different from text and image modality where the generation target can be expressed by natural language or categorized into discrete classes, it is difficult to obtain explicit representation of time series with words or class labels.
Inspired by the widely adopted technique to extract ``basis'' which are the elementary features of time series~\cite{basisformer}, these basis can be utilized as the shared "dictionary" among different domains, each of which encodes different semantic feature for time series.

\subsubsection{Semantic Prototype Module}

Each basis represents certain elementary time series feature like trend and seasonality, that may exist in time series data samples. 
Different individual time series samples are assumed to share the same collection of basis but reflect distinct subset of the collection. 
As a result, each time series gets unique realization of these underlying features, similar to variable weighted allocations to all the basis.
Based on this assumption, a set of latent arrays is introduced as time series prototypes $\bcP\in \mathbb{R}^{N_p\times d}$ for representing cross-domain time series common knowledge, where each prototype vector $\bsp \in \mathbb{R}^{1\times d}$ serves as the representation of a time series basis. 
In practice, the time series prototypes $\bsP$ are initialized with random orthogonal vectors and are frozen afterwards. 

\subsubsection{Prototype Assignment Module~(PAM)} 
Given the assumption that each time series sample corresponds to a distinct allocation of all the basis, the mapping from time series samples to the allocations needs to be established for explicitly identifying important prototypes for each time series instance as well as distinguishing among domains. 
We propose to extract a prototype assignment for each time series instance as the importance weights of each time series on each prototype, and the prototype assignments then serve as conditions for the generation model.

Specifically, each input sequence $\boldsymbol{x}$ is mapped into a weight vector whose dimension equals to the number of prototypes using weight extractor $\phi$, which is a neural network. 
The vector $\phi(\boldsymbol{x})$ represents the weight of each vector inside $\boldsymbol{P}$, and the weights are utilized to modify the attention weight within the cross-attention mechanism so that the predicted noises are only conditioned on the assigned prototypes.
Therefore, sequences from different domains are represented by different $\boldsymbol{m}$ weighted combinations of the shared same set of time series prototypes. 
For ensuring sparsity on prototype assignments, all negative weights are discarded when conducting prototype assignment. 
Formally, the prototype assignments $\boldsymbol{m}$ is extracted with the following formula:
\begin{equation}\label{mask}
    \boldsymbol{m} = \phi(\boldsymbol{x}_0) - \mathbf{I}_{\phi(\boldsymbol{x}_0)<=0} \cdot \infty,
\end{equation}
where $\mathbf{I}_{\cdot<0}$ is the indicator function of negative elements.

\subsection{Domain-Unified Training}\label{sec:trainsam}
Instead of training individual model for each specific dataset, we train one model with data from multiple datasets at the same time for generating different domain data. Here, we treat each dataset as a separate domain. While data from each domain only represent limited fraction of possible data distribution, leveraging data from other domain can help model capture a more diverse time series data distribution. 

Other than taking the unconditional denoising diffusion objective stated in~\cref{uncond_diff}, we employ the conditional denoising objective using $c$ from~\cref{eq:cond} as condition to make use of the encoded semantic context for denoising process, where the conditions are incorporated into the intermediate layers of noise prediction network by spatial attention. 
\begin{align}
    & \bsQ^{(i)} = \bsz^{(i-1)} \cdot \bsW_Q^{(i)}, \\ & \bsK^{(i)} = \bsP \cdot \bsW_K^{(i)},\quad 
    \bsV^{(i)}= \bsP \cdot \bsW_V^{(i)},\quad \\
    &\bsz^{(i)} = \operatorname{FF}(\operatorname{softmax}(\frac{\bsQ^{(i)}\bsK^{(i)T}}{\sqrt{d}}+\boldsymbol{m})\cdot \bsV^{(i)}),
\end{align}
where $\boldsymbol{z}^{(i)} \in \mathbb{R}^{N\times d}$ denotes the output of the $i^{\text{th}}$ last U-Net block. $\bsW_Q^{(i)} \in \mathbb{R}^{d\times d}$, $\bsW_K^{(i)} \in \mathbb{R}^{d\times d}$ and $\bsW_V^{(i)} \in \mathbb{R}^{d\times d}$ are learnable projection matrices applied on the sequence dimension. $\operatorname{FF}$ denotes feed forward layer. The attention output $\boldsymbol{z}^{\text{final}}$ is followed by another feed forward network to produce final block output $\hat{\epsilon}=\operatorname{FF}(\boldsymbol{z}^{\text{final}})$.

With the conditional denoising mechanism described above, the denoising objective using $\epsilon$-parameterization can be written and simplified into:
\begin{align}
    & L_{\textit{cond}} = \mathbb{E}[\|\bep-\hat{\bep}\|^2] \nonumber\\
    &\label{eq:condloss} = \mathbb{E}_{\bsx_0 \in {D^T},{\bep}\sim \mathcal{N}(\mathbf 0,\mathbf I),n}[\|\bep-\bep_{\theta,\bsP}({x}_n,n,\bsm)\|^2].
\end{align}

Due to the imbalance number of training samples across domain, we adopt a re-weight sampling method for making the probability equal for training on samples from each domain. 
Let $N_i$ denote the number of sample sequences in dataset $i$, we set the weight for sampling each sample of this dataset as $w_i=\frac{1}{N_i*|D|}$, such that the probability for sampling sequence from each dataset is balanced. 
The pseudo code for training algorithm is shown at Algorithm~\ref{algo:train}.
\input{src_chapters/3_algo_train.tex}

\subsection{Generation with Domain Prompt}\label{sec:domcond}
To generate time series samples of selected domain after the domain-unified training on multiple datasets, we first extract domain-specific prototype assignments of a small random subset of training samples for the selected domain and group them into a distribution of domain prompt representing the selected domain. Let $K$ denotes the number of selected samples from dataset $i$, the domain prompt is denoted by $\boldsymbol{m}^{D_i}=\{\boldsymbol{m}^{i}_1, ..., \boldsymbol{m}^{i}_K\}$. By constructing the conditioning input, the model generates samples adhere to the selected domain while is not constrained by the general temporal patterns exhibited in the selected samples. 
When the number of expected generated samples is larger than $K$, we use a strategy of repeatedly generate with each assignment in $K$ samples until the number of expected samples is satisfied.

The sampling algorithm is described as Algorithm~\ref{algo:sample}.
\input{src_chapters/3_algo_sample.tex}
\input{src_figtabs/tab_id_gen.tex}
\subsection{Unseen Domain Generation}\label{sec:unseengen}

Since the prototypes provide representations for time series basis, their representation ability is not restricted to the domains in training sets. 
Therefore, they can be utilized to represent unseen domain or datasets.
For any unseen dataset or domain $D_j$ with respect to the training set, we can use extract ``few-shot" samples ${\bsx^j_1,...,\bsx^j_K}$ from the dataset to construct domain prompt $\bsm^{D_j}=\{\boldsymbol{m}^{j}_1, ..., \boldsymbol{m}^{j}_K\}$, and then feed them into the model as condition to generate new samples of the required dataset.

%% file: src_chapters/3_algo_train.tex
\begin{algorithm}
\small
\caption{Training algorithm}
\label{algo:train}
\begin{algorithmic}[1]
\REQUIRE Sequence sample $\boldsymbol{x}$
\ENSURE Network parameters $\phi$ and $\theta$, prototypes $\boldsymbol{P}$
\STATE Initialize prototypes $\boldsymbol{P}$
\REPEAT 
    \STATE Sample $\boldsymbol{x}_0$ from $D^T$
    \STATE Extract prototype assignments $\boldsymbol{m}$ according to~\cref{mask} 
    \STATE Randomly set $\boldsymbol{P}$ as unconditional identifier $\boldsymbol{p}_u$
    \STATE Randomly sample time step $n \sim \mathcal{U}(1,N)$
    \STATE Randomly sample noise $\boldsymbol{\epsilon} \sim \mathcal{N}(\boldsymbol{0},\boldsymbol{I})$
    \STATE Corrupt data $\boldsymbol{x}_n = \sqrt{\bar{\alpha}_n}\boldsymbol{x}_0 + \sqrt{1-\bar{\alpha}_n}\boldsymbol{\epsilon}$
    \STATE Predict step noise with $\tilde{\boldsymbol{\epsilon}} = \tilde{\boldsymbol{\epsilon}}_{\theta,\bsP}(\boldsymbol{x}_n,n,\bo{m})$
    \STATE Compute loss with~\Cref{eq:condloss} and take gradient step.
\UNTIL maximum training step
\end{algorithmic}
\end{algorithm}

%% file: src_chapters/3_algo_sample.tex
\begin{algorithm}
\small
\caption{Sampling with domain prompts}
\label{algo:sample}
\begin{algorithmic}[1]
\REQUIRE K time series prompts $\bo{x}$, prototypes $\bo{P}$
\ENSURE Generated time series samples $\hat{\bo{x}}$
\STATE Extract prototype prompts $\bo{m}$ with $\bo{x}$ according to~\cref{mask}
\STATE Randomly sample noise $\hat{\bo{x}}_N \sim \mathcal{N}(\bo{0},\bo{I})$
\FOR{n from N to 1}
    \STATE Predict step noise with $\tilde{\bo{\epsilon}}_n = \tilde{\bo{\epsilon}}_{\theta,\bsP}(\hat{\bo{x}}_n,n,\bo{m})$
    \STATE Denoise $\hat{\bo{x}}_{n-1} = \frac{\hat{\bo{x}}_n - \sqrt{1-\bar{\alpha}_n}\tilde{\bo{\epsilon}}_n}{\sqrt{\bar{\alpha}_n}}$  
\ENDFOR 
\STATE $\hat{\bo{x}} = \hat{\bo{x}}_0$  
\end{algorithmic}
\end{algorithm}

%% file: src_figtabs/tab_id_gen.tex
{\setlength{\tabcolsep}{4pt}
\begin{table*}[t]
\centering
\small
\begin{tabular}{@{}cccccccc@{}}
\toprule
 &  & \textbf{TimeDP} & \textbf{TimeGAN} & \textbf{GT-GAN} & \textbf{TimeVAE} & \textbf{TimeVQVAE} & \textbf{TimeVQVAE-C} \\ \midrule
\multirow{12}{*}{\rotatebox{90}{Maximum Mean Discrepancy}} & Electricity & $\textbf{0.001}_{\pm   0.001}$ & $0.367_{\pm 0.255}$ & $0.254_{\pm 0.166}$ & $0.577_{\pm 0.006}$ & $0.152_{\pm 0.024}$ & \ul{ $0.002_{\pm 0.001}$} \\
 & Solar & $\textbf{0.041}_{\pm 0.011}$ & $0.628_{\pm 0.053}$ & $0.578_{\pm 0.039}$ & $0.353_{\pm 0.014}$ & $0.437_{\pm 0.020}$ & \ul{ $0.058_{\pm 0.005}$} \\
 & Wind & \ul{ $0.025_{\pm 0.017}$} & $0.213_{\pm 0.017}$ & $0.170_{\pm 0.040}$ & $0.170_{\pm 0.004}$ & $0.131_{\pm 0.014}$ & $\textbf{0.018}_{\pm 0.007}$ \\
 & Traffic & $\textbf{0.083}_{\pm 0.034}$ & $0.567_{\pm 0.057}$ & $0.538_{\pm 0.078}$ & $0.218_{\pm 0.007}$ & $0.213_{\pm 0.016}$ & \ul{ $0.089_{\pm 0.001}$} \\
 & Taxi & $\textbf{0.095}_{\pm 0.023}$ & $0.275_{\pm 0.054}$ & $0.319_{\pm 0.032}$ & $0.139_{\pm 0.007}$ & $0.128_{\pm 0.004}$ & \ul{ $0.109_{\pm 0.014}$} \\
 & Pedestrian & $\textbf{0.044}_{\pm 0.020}$ & $0.090_{\pm 0.030}$ & $0.112_{\pm 0.019}$ & $0.065_{\pm 0.002}$ & $0.067_{\pm 0.007}$ & \ul{ $0.058_{\pm 0.002}$} \\
 & Air & $\textbf{0.011}_{\pm 0.003}$ & $0.120_{\pm 0.045}$ & $0.211_{\pm 0.041}$ & $0.089_{\pm 0.016}$ & \ul{ $0.028_{\pm 0.002}$} & $0.041_{\pm 0.008}$ \\
 & Temperature & $\textbf{0.219}_{\pm 0.022}$ & $0.926_{\pm 0.042}$ & $0.809_{\pm 0.081}$ & $1.002_{\pm 0.014}$ & $0.323_{\pm 0.008}$ & \ul{ $0.259_{\pm 0.043}$} \\
 & Rain & $\textbf{0.057}_{\pm 0.039}$ & $0.329_{\pm 0.285}$ & $0.111_{\pm 0.109}$ & $0.292_{\pm 0.019}$ & \ul{ $0.074_{\pm 0.007}$} & $0.080_{\pm 0.004}$ \\
 & NN5 & $\textbf{0.164}_{\pm 0.010}$ & $0.874_{\pm 0.088}$ & $0.632_{\pm 0.074}$ & $0.821_{\pm 0.061}$ & $0.327_{\pm 0.012}$ & \ul{ $0.243_{\pm 0.041}$} \\
 & Fred-MD & $\textbf{0.002}_{\pm 0.001}$ & $0.043_{\pm 0.021}$ & $0.133_{\pm 0.102}$ & $0.059_{\pm 0.008}$ & $0.008_{\pm 0.002}$ & \ul{ $0.005_{\pm 0.002}$} \\
 & Exchange & $\textbf{0.151}_{\pm 0.024}$ & $0.530_{\pm 0.154}$ & $0.475_{\pm 0.116}$ & $0.543_{\pm 0.149}$ & $0.342_{\pm 0.050}$ & \ul{ $0.233_{\pm 0.107}$} \\ \midrule
\multirow{12}{*}{\rotatebox{90}{K-L}} & Electricity & $\textbf{0.012}_{\pm   0.016}$ & $0.488_{\pm 0.175}$ & $0.407_{\pm 0.079}$ & $0.734_{\pm 0.023}$ & $0.280_{\pm 0.051}$ & \underline{ $0.027_{\pm 0.015}$} \\
 & Solar & $\textbf{0.016}_{\pm 0.005}$ & $0.612_{\pm 0.447}$ & \underline{ $0.120_{\pm 0.041}$} & $0.260_{\pm 0.016}$ & $0.865_{\pm 0.108}$ & $0.234_{\pm 0.062}$ \\
 & Wind & \underline{ $0.152_{\pm 0.034}$} & $1.924_{\pm 1.233}$ & $\textbf{0.107}_{\pm 0.016}$ & $0.484_{\pm 0.015}$ & $0.483_{\pm 0.066}$ & $0.183_{\pm 0.047}$ \\
 & Traffic & $\textbf{0.009}_{\pm 0.003}$ & $1.305_{\pm 0.320}$ & $1.409_{\pm 0.251}$ & $0.211_{\pm 0.014}$ & $0.178_{\pm 0.026}$ & \underline{ $0.016_{\pm 0.003}$} \\
 & Taxi & $\textbf{0.011}_{\pm 0.004}$ & $0.650_{\pm 0.180}$ & $0.950_{\pm 0.197}$ & $0.110_{\pm 0.020}$ & $0.110_{\pm 0.026}$ & \underline{ $0.038_{\pm 0.010}$} \\
 & Pedestrian & $\textbf{0.014}_{\pm 0.010}$ & $0.417_{\pm 0.181}$ & $0.411_{\pm 0.096}$ & $0.065_{\pm 0.005}$ & $0.405_{\pm 0.051}$ & \underline{ $0.039_{\pm 0.008}$} \\
 & Air & $\textbf{0.027}_{\pm 0.016}$ & $0.348_{\pm 0.093}$ & $0.578_{\pm 0.049}$ & $0.164_{\pm 0.012}$ & \underline{ $0.054_{\pm 0.012}$} & $0.093_{\pm 0.025}$ \\
 & Temperature & $\textbf{0.171}_{\pm 0.073}$ & $8.892_{\pm 2.681}$ & $3.174_{\pm 2.685}$ & $2.183_{\pm 0.110}$ & $0.735_{\pm 0.066}$ & \underline{ $0.379_{\pm 0.110}$} \\
 & Rain & $\textbf{0.013}_{\pm 0.012}$ & $0.506_{\pm 0.174}$ & $0.432_{\pm 0.099}$ & $0.160_{\pm 0.022}$ & \underline{ $0.047_{\pm 0.018}$} & $0.065_{\pm 0.018}$ \\
 & NN5 & $\textbf{0.054}_{\pm 0.014}$ & $4.928_{\pm 4.112}$ & $1.386_{\pm 0.520}$ & $1.337_{\pm 0.220}$ & $1.063_{\pm 0.274}$ & \underline{ $0.220_{\pm 0.151}$} \\
 & Fred-MD & $\textbf{0.203}_{\pm 0.035}$ & $0.512_{\pm 0.290}$ & $0.380_{\pm 0.070}$ & \underline{ $0.346_{\pm 0.041}$} & $0.831_{\pm 0.077}$ & $1.118_{\pm 0.276}$ \\
 & Exchange & $\textbf{1.866}_{\pm 0.132}$ & $8.861_{\pm 3.397}$ & $7.201_{\pm 4.380}$ & $10.404_{\pm 1.434}$ & \underline{$5.052_{\pm 1.385}$} & $8.475_{\pm 3.056}$ \\ \bottomrule
\end{tabular}
\caption{Maximum mean discrepancy~(MMD) and K-L divergence~(K-L) of in-domain generation for sequence length 168. Best results are highlighted in bold face and second best results are underlined.}
\label{tab:indomain}
\end{table*}
}

%% file: src_chapters/4_exps.tex
\section{Experiments}
In this section, we provide empirical experiment results for our method using multiple real-world datasets. The experiment goal is to investigate on the following research questions: (a) How good is the quality of prompted generation on trained domain? (b) Can the learned prototype help the model generalize to unseen domains? 
\subsection{Experiment Settings}
\paragraph{Datasets}
The experiments are conducted on 12 datasets across four time series domains: Electricity, Solar and Wind from the energy domain; Traffic, Taxi and Pedestrian from the transport domain, Air Quality, Temperature and Rain from the nature domain; NN5, Fred-MD and Exchange from the economic domain. All datasets are obtained by GluonTS package and Monash Time Series Forecasting Repository. We pre-process all datasets into non-overlapping uni-variate sequence slices with length in $\{24, 96, 168, 336\}$. More details on datasets are described in the technical appendix.

\paragraph{Baselines}
We compare our generation results with several representative state-of-the-art time series generation methods, including TimeGAN~\citep{timegan}, GT-GAN~\citep{gtgan}, TimeVAE~\citep{timevae}, and TimeVQVAE~\citep{timevq}. Note that in their original implementation, these methods are trained with single dataset from a single domain for each time. To align their implementations with our problem setting and method, we train these models with the multi-domain dataset which mix all of the twelve datasets together. Note that the TimeVQVAE baseline has a class-conditioned variant, which we considered as a conditional time series generation baseline that is able to sample by domain label after being trained with the multi-domain dataset.
\input{src_figtabs/tab_zero_168.tex}
\paragraph{Implementation Details}
We follow common practice of diffusion models to utilize a U-Net architecture for our denoising model. The architecture details are discussed in the technical appendix. The number of prototypes are set to 16 for all the main evaluations. Models for each sequence length are trained for $50,000$ steps using a batch size of 128 and a learning rate of $1\times 10^{-4}$ with $1,000$ warm-up steps. For all baselines, we take the implementations and recommended hyper-parameter settings from their public codes.

\paragraph{Evaluation Metrics}
The major consideration for evaluating performance of time series generation methods are two fold: (1) the similarity between real and synthetic time series distributions; (2) the internal temporal dependency within each instance. In light of these two we apply the following three metrics for evaluating generation performance: (1) \textbf{Maximum Mean Discrepancy~(MMD)} which maps the data points into a high-dimensional feature space with a kernel function and then compare the means of the two data distribution; (2) \textbf{Kullback-Leibler divergence~(K-L)} which is the measure of how one probability distribution diverges from a second, reference probability distribution, indicating the distribution difference on variable level; (3) \textbf{Marginal distribution difference~(MDD)} which calculates the empirical histogram for each time step and calculate the average absolute difference of real and synthetic data across bins. 

\subsection{Results}
\subsubsection{Evaluation of Generation Quality}
In this experiment, we evaluate the time series generation performance of our model on all 12 datasets across four domains, comparing with the baselines on MMD, K-L and MDD. The proposed model and baseline models are evaluated after trained with the multi-domain dataset. 
While the baselines are not designed for learning time series from multiple domains simultaneously, we modify them to align with our setting that treating the multi-domain dataset as a whole. Additionally, while TimeVQVAE can be trained in a class-conditional manner, we also test on this variant, using the source of data domain as class label to obtain a cross-domain version, as shown in column ``TimeVQVAE-C". Each run is repeated 5 times with different random seeds. The average and the standard deviation of MMD and K-L results for the sequence length of 168 are shown in~\Cref{tab:indomain}. The MDD results and other results for different sequence length settings are reported in the appendix.

As shown in~\Cref{tab:indomain}, our approach achieves the best results on most of the twelve in-domain datasets, demonstrating superior performance on generating new time series samples that have the closest distribution to real dataset samples, both jointly and marginally. Most second best scores are obtained with the class-conditional version of TimeVQVAE model. The results indicates that, simply mixing datasets as the multi-domain training approach generally fails to generate samples that are close to real data in distribution, majorly due to the diverse intrinsic pattern among different domains. Although using dataset labels helps greatly on the training with mixed data as shown with TimeVQVAE-C, our model outperforms it without being explicitly supervised with class labels, indicating strong representation disentanglement capability. The results for other sequence length settings are included in the technical appendix. The results show that TimeDP consistently demonstrate the best generation performance on both long and short sequence length scenarios.
\input{src_figtabs/tab_abla_large.tex}
\subsubsection{Evaluation of Unseen Domain Time Series Synthesis}
In this experiment, we evaluate the synthesis performance on unseen domain, where datasets that are not included in the training data. We randomly select a few samples from the new datasets as demonstrations for all models, and ensure a test set that does not overlap with these samples. For TimeDP, these samples are used to extract domain prompts. For baseline methods that are not designed to generate following prompts or can not generate with unseen class labels, we evaluate their unconditional generation outputs. The results for these baseline models are labeled with ``unconditional" at the table. 
Additionally, we conduct a fine-tuning on all baseline models to evaluate their few-shot learning setting. The fine-tuned model are labeled with ``Fine-tuned" at the table. We use $3,10$ and $100$ samples as ``prompt" and fine-tuning data in this experiment. The MMD and K-L results of our model and baselines for the generation sequence length 168 are shown in~\Cref{tab:zerolargenew}.

\Cref{tab:zerolargenew} shows that our model demonstrates robust zero-shot time series synthesis capability, obtaining the best general MMD and K-L scores compared to baselines. The fine-tuned models do not show consistent performance improvement against the unconditional model, indicating that in the low data regime, fine-tuning is not effective for fitting the data distribution. On the contrary, our model is able to infer the unseen domain distribution using domain prompts without additional tuning, and the performance improves as the number of few-shot samples grows, showing strong unseen domain generation capability.

\subsubsection{Ablation Study}
In this section, we evaluate the sensitivity of PAM design by modifying the number of prototypes, and evaluate its effectiveness by removing it and using $\phi({\bsx})$ as domain prompt instead, which is shown with ``-PAM''. We also test the unconditional version of model, labeled with ``-Prompt''. \Cref{tab:ablation} shows the results on MMD and K-L scores, and MDD scores are shown in the appendix. Our model's performance remains consistent when the number is large enough. Removing PAM and removing conditioning mechanism both lead to great drop in MMD score while maintaining stable K-L scores, indicating that diffusion model backbone is strong enough to capture marginal distribution while PAM and domain prompts are essential in capturing sequence-wise time series distribution. 

%% file: src_figtabs/tab_zero_168.tex
{\setlength{\tabcolsep}{4pt}
\begin{table*}[t]
\centering
\small
\begin{tabular}{@{}ccccccccc@{}}
\toprule
 &  &  & \multicolumn{3}{c}{Stock} & \multicolumn{3}{c}{Web} \\ \cmidrule(lr){4-6} \cmidrule(lr){7-9}
 &  & \# Shots & 3 & 10 & 100 & 3 & 10 & 100 \\ \midrule
\multirow{9}{*}{\rotatebox{90}{\makecell{Maximum Mean\\Discrepancy}}} & \multirow{4}{*}{Unconditional} & TimeGAN & $0.517_{\pm 0.030}$ & $0.823_{\pm 0.042}$ & $0.944_{\pm 0.075}$ & $0.197_{\pm 0.027}$ & $0.089_{\pm 0.045}$ & $0.126_{\pm 0.104}$ \\
 &  & GT-GAN & $0.466_{\pm 0.050}$ & $0.790_{\pm 0.048}$ & $0.692_{\pm 0.074}$ & $0.109_{\pm 0.027}$ & $0.117_{\pm 0.012}$ & $0.112_{\pm 0.014}$ \\
 &  & TimeVAE & $0.351_{\pm 0.009}$ & $0.870_{\pm 0.022}$ & $0.898_{\pm 0.025}$ & $0.203_{\pm 0.004}$ & $0.146_{\pm 0.006}$ & $0.099_{\pm 0.006}$ \\
 &  & TimeVQVAE & \ul{ $0.105_{\pm 0.005}$} & \ul{ $0.111_{\pm 0.001}$} & \ul{ $0.096_{\pm 0.004}$} & \ul{ $0.078_{\pm 0.007}$} & $\textbf{0.063}_{\pm 0.009}$ & $0.062_{\pm 0.006}$ \\ \cmidrule(l){2-9} 
 & \multirow{4}{*}{Fine-tuned} & TimeGAN & $0.124_{\pm 0.081}$ & $0.144_{\pm 0.034}$ & $0.136_{\pm 0.093}$ & $0.157_{\pm 0.034}$ & $0.182_{\pm 0.054}$ & \ul{ $0.055_{\pm 0.018}$} \\
 &  & GT-GAN & $0.139_{\pm 0.076}$ & $0.161_{\pm 0.113}$ & $0.261_{\pm 0.132}$ & $0.101_{\pm 0.053}$ & $0.191_{\pm 0.037}$ & $0.230_{\pm 0.068}$ \\
 &  & TimeVAE & $0.348_{\pm 0.007}$ & $0.790_{\pm 0.028}$ & $0.761_{\pm 0.032}$ & $0.199_{\pm 0.013}$ & $0.160_{\pm 0.045}$ & $0.118_{\pm 0.019}$ \\
 &  & TimeVQVAE & $0.205_{\pm 0.089}$ & $0.176_{\pm 0.095}$ & $0.127_{\pm 0.050}$ & $0.278_{\pm 0.226}$ & $0.198_{\pm 0.077}$ & $0.085_{\pm 0.060}$ \\ \cmidrule(l){2-9} 
 & Prompted & DP-Diff & $\textbf{0.048}_{\pm 0.014}$ & $\textbf{0.050}_{\pm 0.012}$ & $\textbf{0.014}_{\pm 0.005}$ & $\textbf{0.068}_{\pm 0.035}$ & \ul{ $0.065_{\pm 0.014}$} & $\textbf{0.021}_{\pm 0.004}$ \\ \midrule
\multirow{9}{*}{\rotatebox{90}{K-L}} & \multirow{4}{*}{Unconditional} & TimeGAN & $5.077_{\pm 1.314}$ & $5.213_{\pm 1.211}$ & $3.717_{\pm 1.563}$ & $1.485_{\pm 0.766}$ & $0.432_{\pm 0.375}$ & $0.485_{\pm 0.340}$ \\
 &  & GT-GAN & $2.010_{\pm 0.302}$ & $2.086_{\pm 0.294}$ & $1.541_{\pm 0.120}$ & $0.645_{\pm 0.123}$ & $1.351_{\pm 0.291}$ & $1.626_{\pm 0.274}$ \\
 &  & TimeVAE & $1.872_{\pm 0.094}$ & $2.175_{\pm 0.106}$ & $1.666_{\pm 0.075}$ & $0.345_{\pm 0.021}$ & $\textbf{0.056}_{\pm 0.007}$ & \ul{ $0.104_{\pm 0.007}$} \\
 &  & TimeVQVAE & $\textbf{0.840}_{\pm 0.117}$ &$\textbf{0.458}_{\pm 0.084}$ & $0.692_{\pm 0.125}$ & $\textbf{0.175}_{\pm 0.035}$ & $0.448_{\pm 0.034}$ & $0.579_{\pm 0.045}$ \\ \cmidrule(l){2-9} 
 & \multirow{4}{*}{Fine-tuned} & TimeGAN & $3.164_{\pm 0.361}$ & $1.844_{\pm 0.684}$ & $1.304_{\pm 0.450}$ & $1.739_{\pm 0.662}$ & $0.606_{\pm 0.119}$ & $0.357_{\pm 0.140}$ \\
 &  & GT-GAN & $1.183_{\pm 0.266}$ & \ul{ $0.482_{\pm 0.253}$} & \ul{ $0.681_{\pm 0.292}$} & $0.588_{\pm 0.313}$ & $1.230_{\pm 0.248}$ & $1.640_{\pm 0.553}$ \\
 &  & TimeVAE & $1.331_{\pm 0.269}$ & $1.855_{\pm 0.170}$ & $1.344_{\pm 0.080}$ & $0.309_{\pm 0.050}$ & \ul{ $0.058_{\pm 0.010}$} & $0.482_{\pm 0.097}$ \\
 &  & TimeVQVAE & \ul{ $1.139_{\pm 0.416}$} & $0.614_{\pm 0.303}$ & $0.864_{\pm 0.688}$ & $1.176_{\pm 1.338}$ & $0.540_{\pm 0.318}$ & $0.383_{\pm 0.295}$ \\ \cmidrule(l){2-9}
 & Prompted & DP-Diff & $1.613_{\pm 0.065}$ & $0.721_{\pm 0.041}$ & $\textbf{0.090}_{\pm 0.037}$ & \ul{ $0.245_{\pm 0.027}$} & $0.254_{\pm 0.030}$ &$\textbf{0.091}_{\pm 0.010}$ \\ \midrule
 \multirow{9}{*}{\rotatebox{90}{MDD}} & \multirow{4}{*}{Unconditional} & TimeGAN & $0.837_{\pm 0.014}$ & $0.803_{\pm 0.007}$ & $0.790_{\pm 0.015}$ & $0.084_{\pm 0.008}$ & \ul{ $0.054_{\pm 0.013}$} & $0.062_{\pm 0.022}$ \\
 &  & GT-GAN & $0.795_{\pm 0.007}$ & $0.804_{\pm 0.002}$ & $0.756_{\pm 0.006}$ & $0.050_{\pm 0.005}$ & $0.064_{\pm 0.002}$ & $0.059_{\pm 0.002}$ \\
 &  & TimeVAE & $0.814_{\pm 0.003}$ & $0.858_{\pm 0.003}$ & $0.859_{\pm 0.003}$ & $0.076_{\pm 0.002}$ & $0.057_{\pm 0.002}$ & \ul{ $0.045_{\pm 0.001}$} \\
 &  & TimeVQVAE & $\textbf{0.711}_{\pm 0.005}$ & \ul{ $0.702_{\pm 0.003}$} & \ul{ $0.706_{\pm 0.006}$} & \ul{ $0.047_{\pm 0.005}$} & $0.072_{\pm 0.003}$ & $0.073_{\pm 0.002}$ \\ \cmidrule(l){2-9} 
 & \multirow{4}{*}{Fine-tuned} & TimeGAN & $0.750_{\pm 0.031}$ & $0.770_{\pm 0.046}$ & $0.760_{\pm 0.038}$ & $0.066_{\pm 0.022}$ & $0.079_{\pm 0.003}$ & $0.062_{\pm 0.012}$ \\
 &  & GT-GAN & \ul{ $0.717_{\pm 0.026}$} & $0.711_{\pm 0.032}$ & $0.731_{\pm 0.018}$ & $0.048_{\pm 0.010}$ & $0.072_{\pm 0.010}$ & $0.080_{\pm 0.010}$ \\
 &  & TimeVAE & $0.801_{\pm 0.012}$ & $0.847_{\pm 0.005}$ & $0.840_{\pm 0.006}$ & $0.077_{\pm 0.002}$ & $0.056_{\pm 0.004}$ & $0.071_{\pm 0.006}$ \\
 &  & TimeVQVAE & $0.741_{\pm 0.022}$ & $0.718_{\pm 0.029}$ & $0.708_{\pm 0.024}$ & $0.076_{\pm 0.046}$ & $0.087_{\pm 0.015}$ & $0.062_{\pm 0.025}$ \\ \cmidrule(l){2-9} 
 & Prompted & DP-Diff & $0.720_{\pm 0.012}$ & $\textbf{0.690}_{\pm 0.009}$ & $\textbf{0.653}_{\pm 0.002}$ & $\textbf{0.042}_{\pm 0.006}$ & $\textbf{0.051}_{\pm 0.005}$ & $\textbf{0.038}_{\pm 0.002}$ \\ \bottomrule
\end{tabular}
\caption{Maximum mean discrepancy~(MMD), K-L divergence~(K-L) and Marginal distribution difference~(MDD) in unseen domain settings of Stock and Web datasets for generation sequence length 168. Best results are highlighted in bold face.}
\label{tab:zerolargenew}
\end{table*}
}

%% file: src_figtabs/tab_abla_large.tex
{\setlength{\tabcolsep}{4pt}
\begin{table*}[t]
\centering
\small
\begin{tabular}{@{}ccccccccc@{}}
\toprule
 &  & \multicolumn{5}{c}{TimeDP} & \multirow{2}{*}{- PAM} & \multirow{2}{*}{- Prompt} \\ \cmidrule(r){3-7}
 & \# Prototypes & 4 & 8 & 16 & 32 & 64 &  &  \\ \midrule
\multirow{13}{*}{\rotatebox{90}{Maximum Mean Discrepancy}} & Electricity & $0.004_{\pm 0.002}$ & $0.001_{\pm 0.001}$ & \ul{ $0.001_{\pm 0.001}$} & $0.001_{\pm 0.001}$ & $\textbf{0.001}_{\pm 0.001}$ & $0.004_{\pm 0.002}$ & $0.016_{\pm 0.015}$ \\
 & Solar & $0.061_{\pm 0.025}$ & $0.041_{\pm 0.018}$ & $0.041_{\pm 0.011}$ & \ul{ $0.037_{\pm 0.007}$} & $\textbf{0.037}_{\pm 0.006}$ & $0.545_{\pm 0.030}$ & $0.520_{\pm 0.029}$ \\
 & Wind & $0.029_{\pm 0.017}$ & $0.027_{\pm 0.017}$ & \ul{ $0.025_{\pm 0.017}$} & $0.025_{\pm 0.017}$ & $\textbf{0.025}_{\pm 0.016}$ & $0.134_{\pm 0.028}$ & $0.074_{\pm 0.047}$ \\
 & Traffic & $0.101_{\pm 0.025}$ & \ul{ $0.077_{\pm 0.018}$} & $0.083_{\pm 0.034}$ & $0.081_{\pm 0.023}$ & $\textbf{0.067}_{\pm 0.005}$ & $0.270_{\pm 0.015}$ & $0.288_{\pm 0.020}$ \\
 & Taxi & $0.111_{\pm 0.021}$ & $0.095_{\pm 0.013}$ & \ul{ $0.095_{\pm 0.023}$} & $0.095_{\pm 0.020}$ & $\textbf{0.093}_{\pm 0.016}$ & $0.149_{\pm 0.026}$ & $0.140_{\pm 0.014}$ \\
 & Pedestrian & $0.049_{\pm 0.021}$ & $\textbf{0.040}_{\pm 0.010}$ & $0.044_{\pm 0.020}$ & $0.044_{\pm 0.016}$ & \ul{ $0.041_{\pm 0.016}$} & $0.052_{\pm 0.009}$ & $0.057_{\pm 0.007}$ \\
 & Air & $0.016_{\pm 0.003}$ & $0.011_{\pm 0.001}$ & \ul{ $0.011_{\pm 0.003}$} & $0.012_{\pm 0.003}$ & $\textbf{0.011}_{\pm 0.002}$ & $0.061_{\pm 0.031}$ & $0.017_{\pm 0.007}$ \\
 & Temperature & $0.221_{\pm 0.034}$ & \ul{ $0.209_{\pm 0.014}$} & $0.219_{\pm 0.022}$ & $0.221_{\pm 0.023}$ & $0.224_{\pm 0.027}$ & $0.354_{\pm 0.011}$ & $\textbf{0.179}_{\pm 0.043}$ \\
 & Rain & $0.056_{\pm 0.020}$ & \ul{ $0.047_{\pm 0.020}$} & $0.057_{\pm 0.039}$ & $0.051_{\pm 0.030}$ & $0.057_{\pm 0.057}$ & $0.056_{\pm 0.046}$ & $\textbf{0.043}_{\pm 0.018}$ \\
 & NN5 & $0.198_{\pm 0.059}$ & \ul{ $0.157_{\pm 0.003}$} & $0.164_{\pm 0.010}$ & $\textbf{0.155}_{\pm 0.008}$ & $0.158_{\pm 0.005}$ & $0.348_{\pm 0.021}$ & $0.257_{\pm 0.021}$ \\
 & Fred-MD & $\textbf{0.001}_{\pm 0.000}$ & \ul{ $0.002_{\pm 0.001}$} & $0.002_{\pm 0.001}$ & $0.002_{\pm 0.001}$ & $0.002_{\pm 0.001}$ & $0.338_{\pm 0.114}$ & $0.005_{\pm 0.005}$ \\
 & Exchange & \ul{ $0.146_{\pm 0.031}$} & $0.146_{\pm 0.022}$ & $0.151_{\pm 0.024}$ & $0.150_{\pm 0.024}$ & $0.152_{\pm 0.023}$ & $1.059_{\pm 0.008}$ & $\textbf{0.117}_{\pm 0.028}$ \\ \cmidrule(l){2-9} 
 & Average & 0.083 & \textbf{0.071} & 0.074 & 0.073 & \ul{ 0.072} & 0.281 & 0.143 \\ \midrule
\multirow{12}{*}{\rotatebox{90}{K-L}} & Electricity & $0.003_{\pm 0.002}$ & \ul{ $0.011_{\pm 0.014}$} & $0.012_{\pm 0.016}$ & $0.014_{\pm 0.019}$ & $0.012_{\pm 0.018}$ & $\textbf{0.011}_{\pm 0.009}$ & $0.013_{\pm 0.013}$ \\
 & Solar & $0.028_{\pm 0.021}$ & $0.016_{\pm 0.007}$ & $0.016_{\pm 0.005}$ & $\textbf{0.013}_{\pm 0.002}$ & \ul{ $0.014_{\pm 0.005}$} & $0.019_{\pm 0.004}$ & $0.042_{\pm 0.036}$ \\
 & Wind & \ul{ $0.151_{\pm 0.027}$} & $0.153_{\pm 0.044}$ & $0.152_{\pm 0.034}$ & $0.156_{\pm 0.037}$ & $0.155_{\pm 0.032}$ & $\textbf{0.072}_{\pm 0.010}$ & $0.156_{\pm 0.073}$ \\
 & Traffic & $0.024_{\pm 0.030}$ & $\textbf{0.007}_{\pm 0.003}$ & \ul{ $0.009_{\pm 0.003}$} & $0.012_{\pm 0.009}$ & $0.011_{\pm 0.004}$ & $0.014_{\pm 0.013}$ & $0.047_{\pm 0.027}$ \\
 & Taxi & $0.051_{\pm 0.033}$ & \ul{ $0.008_{\pm 0.003}$} & $0.011_{\pm 0.004}$ & $0.012_{\pm 0.010}$ & $0.009_{\pm 0.005}$ & $\textbf{0.003}_{\pm 0.001}$ & $0.032_{\pm 0.022}$ \\
 & Pedestrian & $0.031_{\pm 0.028}$ & $\textbf{0.010}_{\pm 0.008}$ & $0.014_{\pm 0.010}$ & $0.014_{\pm 0.009}$ & \ul{ $0.012_{\pm 0.008}$} & $0.021_{\pm 0.010}$ & $0.041_{\pm 0.047}$ \\
 & Air & $0.047_{\pm 0.027}$ & \ul{ $0.024_{\pm 0.008}$} & $0.027_{\pm 0.016}$ & $0.027_{\pm 0.010}$ & $0.025_{\pm 0.012}$ & $\textbf{0.020}_{\pm 0.005}$ & $0.036_{\pm 0.017}$ \\
 & Temperature & $0.304_{\pm 0.079}$ & $0.176_{\pm 0.061}$ & $0.171_{\pm 0.073}$ & $0.179_{\pm 0.087}$ & \ul{ $0.166_{\pm 0.074}$} & $\textbf{0.132}_{\pm 0.009}$ & $0.168_{\pm 0.096}$ \\
 & Rain & $0.033_{\pm 0.009}$ & \ul{ $0.008_{\pm 0.004}$} & $0.013_{\pm 0.012}$ & $\textbf{0.008}_{\pm 0.005}$ & $0.013_{\pm 0.008}$ & $0.009_{\pm 0.008}$ & $0.009_{\pm 0.004}$ \\
 & NN5 & $0.217_{\pm 0.223}$ & $0.050_{\pm 0.021}$ & $0.054_{\pm 0.014}$ & $\textbf{0.045}_{\pm 0.007}$ & \ul{ $0.046_{\pm 0.006}$} & $0.049_{\pm 0.016}$ & $0.073_{\pm 0.031}$ \\
 & Fred-MD & $\textbf{0.188}_{\pm 0.012}$ & \ul{ $0.190_{\pm 0.013}$} & $0.203_{\pm 0.035}$ & $0.216_{\pm 0.052}$ & $0.201_{\pm 0.024}$ & $0.196_{\pm 0.008}$ & $0.196_{\pm 0.016}$ \\
 & Exchange & $1.863_{\pm 0.223}$ & $1.828_{\pm 0.131}$ & $1.866_{\pm 0.132}$ & $2.313_{\pm 1.007}$ & $1.875_{\pm 0.235}$ & \ul{ $1.623_{\pm 0.146}$} & $\textbf{1.506}_{\pm 0.086}$ \\ \cmidrule(l){2-9} 
 & Average & 0.245 & 0.207 & 0.212 & 0.251 & 0.212 & \textbf{0.181} & \ul{ 0.193} \\ \bottomrule
\end{tabular}

\caption{Maximum mean discrepancy~(MMD) and K-L divergence~(K-L) results for ablation study on generation sequence length 168. Best results are highlighted in bold face. Second best results are underlined.}
\label{tab:ablation}
\end{table*}
 }

%% file: src_chapters/5_conclusion.tex
\section{Conclusion}
In this paper, we propose a multi-domain time series diffusion model with domain prompts, named \textbf{TimeDP}. 
TimeDP learns prototypes to represent time series basis, serving as ``word" with time series semantics. 
A prototype assignment module is applied to extract the domain specific prototype weights, for learning domain prompts as generation condition.
During sampling, we construct ``domain prompt" with few-shot samples from the target domain and use the domain prompts as condition to generate time series samples.
Experiments demonstrate that our method outperforms baselines to provide the state-of-the-art in-domain generation quality and strong unseen domain generation capability.

%% file: src_chapters/appendix.tex
\section{Technical Appendix}
\subsection{Model Details}
\subsubsection{Architecture}
The denoising network of the diffusion model used in this paper is built with U-Net structure, with 4 up/down sampling blocks. Each up/down sampling block consists of 2 residual blocks and 1 cross-attention block, where each residual block contains two 1-D convolution layers and each cross-attention block uses 1-D convolution as input/output projection layers and uses 8 attention heads. There is a middle block containing two residual blocks and one attention blocks, placed after the down sampling blocks and before the up sampling blocks. There are also additional input and output blocks, each with one 1-D convolution layer. We use SiLU as the non-linear activation function in this module.

For the prototype assignment module, we use two 1-D convolution layers as feature extractor. Then there are another two 1-D convolution layers with residual connection, followed by a linear projection layer, as assignment generator producing final prototype assignment. All non-linear activation functions in this module are chosen to be ReLU.

\subsubsection{Description on Datasets}
In this section, we provide descriptions into the datasets used in model training this paper:
\begin{itemize}
    \item \textbf{Electricity}. This dataset represents the hourly electricity consumption of 321 clients from 2012 to 2014 in kilowatt(kW). It was originally extracted from UCI.
    \item \textbf{Solar}. This dataset contains 137 time series representing the solar power production recorded every 1 hour in the state of Alabama in 2006.
    \item \textbf{Wind}. This dataset contains a single very long daily time series representing the wind power production in MW recorded per every 4 seconds starting from 01/08/2019. It was downloaded from the Australian Energy Market Operator (AEMO) online platform. 
    \item \textbf{Traffic}. This dataset contains 15 months worth of daily data (440 daily records) that describes the occupancy rate, between 0 and 1, of different car lanes of the San Francisco bay area freeways across time. 
    \item \textbf{Taxi}. This dataset contains spatio-temporal traffic time series of New York taxi rides taken at 1214 locations every 30 minutes in the months of January 2015 and January 2016.
    \item \textbf{Pedestrian}. This dataset contains hourly pedestrian counts captured from 66 sensors in Melbourne city starting from May 2009. The original dataset is regularly updated when a new set of observations become available. The dataset uploaded here contains pedestrian counts up to 2020-04-30.
    \item \textbf{Air Quality}. This dataset was used in the KDD Cup 2018 forecasting competition. It contains long hourly time series representing the air quality levels in 59 stations in 2 cities: Beijing (35 stations) and London (24 stations) from 01/01/2017 to 31/03/2018. The air quality level is represented in multiple measurements such as PM2.5, PM10, NO2, CO, O3 and SO2. The dataset uploaded here contains 270 hourly time series which have been categorized using city, station name and air quality measurement. The original dataset contains missing values. The leading missing values of a given series were replaced by zeros and the remaining missing values were replaced by carrying forward the corresponding last observations (LOCF method). 
    \item \textbf{Temperature}. This dataset contains 32072 daily time series showing the temperature observations and rain forecasts, gathered by the Australian Bureau of Meteorology for 422 weather stations across Australia, between 02/05/2015 and 26/04/2017. The original dataset contains missing values and they have been simply replaced by zeros. We extracted the mean temperature column here.
    \item \textbf{Rain}. This data set comes from same source as Temperature dataset, extracted the rain column.
    \item \textbf{NN5}. This dataset was used in the NN5 forecasting competition. It contains 111 time series from the banking domain. The goal is predicting the daily cash withdrawals from ATMs in UK. The original dataset contains missing values. A missing value on a particular day is replaced by the median across all the same days of the week along the whole series. 
    \item \textbf{Fred-MD}. This dataset contains 107 monthly time series showing a set of macro-economic indicators from the Federal Reserve Bank. It was extracted from the FRED-MD database. The series are differentiated and log-transformed as suggested in the literature. 
    \item \textbf{Exchange}. This dataset contains daily exchange rate between 8 currencies.
\end{itemize}

More over, we also describe the dataset used in unseen data generation experiment:
\begin{itemize}
    \item \textbf{Stock}. This dataset contains daily stock price of symbol GOOG, listed in NASDAQ.
    \item \textbf{Web}. This dataset was used in the Kaggle Wikipedia Web Traffic forecasting competition. It contains 145063 daily time series representing the number of hits or web traffic for a set of Wikipedia pages from 2015-07-01 to 2017-09-10. The original dataset contains missing values. They have been simply replaced by zeros. 
\end{itemize}

\subsection{Additional Results for Generation Quality Experiment}
The additional marginal distribution distance score results for the generation quality experiment are shown in~\Cref{tab:indomain_mdd}. \Cref{tab:indomain_mdd} shows that TimeDP gets overall best result on the MDD metrics, obtaining the best score for 10 out of 12 datasets.

\Cref{tab:indomain_24},~\Cref{tab:indomain_96} and~\Cref{tab:indomain_336} show results for sequence lengths 24, 96 and 336 respectively. These results show that TimeDP consistently outperforms baselines when generating time series with different sequence lengths, verifying the robustness of TimeDP.

\input{src_figtabs/tab_app_id_mdd.tex}

\input{src_figtabs/tab_app_ablamdd.tex}

\input{src_figtabs/tab_app_id_24.tex}

\input{src_figtabs/tab_app_id_96.tex}

\input{src_figtabs/tab_app_id_336.tex}

\subsection{Marginal Distribution Distance Score for Ablation Study}
The additional marginal distribution distance score results for the ablation study experiment are shown in \Cref{tab:ablation_mdd}. From \Cref{tab:ablation_mdd} we can observe that the marginal distribution distance score performance of TimeDP is generally stable.

\subsection{Visualization Analysis on Domain Prompts}
In this section, we analyze the correlation among domain prompts through visualization. 

\Cref{fig:dp_heat} provides a heatmap for domain prompts, where each row represents one sequence sample of the dataset and each column shows the weights for each prototype. In~\Cref{fig:dp_heat}, we can observe that samples of the same dataset are assigned similar weights on each prototype, and the distribution of weights among different datasets are diverse.

\Cref{fig:dp_dataset_heat} and \Cref{fig:dp_domain_heat} provide t-SNE analysis on the similarity of prompts among domain. We can observe that the model has learned to differentiate among domain with domain prompts.

\subsection{Visualization Analysis on Time Series Semantic Prototypes}
To better understand the role of time series prototypes in TimeDP, we visualize the semantics of each time series prototype by prompting TimeDP model with one-hot vector such that the generation process is conditioned on only one prototype for each generation.~\Cref{fig:dp_semantic} shows the visualization results. We can observe that each prototype learns distinct time series semantic. For example, prototype 0, 4, 5, 6 encodes different seasonality patterns wile prototype 1, 2, 3, 10 encodes strong trend patterns.
\input{src_figtabs/fig_prompt_sem.tex}

\subsection{Visualization on Unseen Domain Generated Time Series}

\Cref{fig:stock10_shots} and~\Cref{fig:web10_shots} show the few-shot prompting experiment results. The plots show that TimeDP can generate new time series samples that are the most similar to the real dataset samples without any finetuning, while the baselines models that are finetuned with the few-shot samples fail to generate realistic time series sequences.

\input{src_figtabs/fig_prompt_heatmap.tex}

\input{src_figtabs/fig_data_tsne.tex}

\input{src_figtabs/fig_domain_tsne.tex}

\input{src_figtabs/fig_web_zs.tex}

\input{src_figtabs/fig_stock_zs.tex}

%% file: src_figtabs/tab_app_id_mdd.tex
{\setlength{\tabcolsep}{4pt}
\begin{table*}[t]
\centering
\small
\begin{tabular}{@{}cccccccc@{}}
\toprule
 &  & \textbf{TimeDP} & \textbf{TimeGAN} & \textbf{GT-GAN} & \textbf{TimeVAE} & \textbf{TimeVQVAE} & \textbf{TimeVQVAE-C} \\ \midrule
\multirow{12}{*}{\rotatebox{90}{Marginal Distribution Distance}} & Electricity & $\textbf{0.005}_{\pm   0.002}$ & $0.075_{\pm 0.035}$ & $0.047_{\pm 0.008}$ & $0.098_{\pm 0.003}$ & $0.067_{\pm 0.004}$ & \ul{$0.005_{\pm 0.001}$} \\
 & Solar & \ul{$56.414_{\pm 21.890}$} & $70.334_{\pm 11.928}$ & $83.855_{\pm 3.100}$ & $\textbf{16.721}_{\pm 0.041}$ & $57.401_{\pm 0.041}$ & $59.043_{\pm 0.440}$ \\
 & Wind & $\textbf{0.084}_{\pm 0.009}$ & $0.226_{\pm 0.061}$ & $0.138_{\pm 0.015}$ & $0.201_{\pm 0.004}$ & $0.159_{\pm 0.011}$ & \ul{$0.085_{\pm 0.008}$} \\
 & Traffic & $\textbf{0.049}_{\pm 0.004}$ & $0.149_{\pm 0.018}$ & $0.153_{\pm 0.001}$ & $0.110_{\pm 0.001}$ & $0.119_{\pm 0.005}$ & \ul{$0.053_{\pm 0.002}$} \\
 & Taxi & $\textbf{0.081}_{\pm 0.008}$ & $0.104_{\pm 0.010}$ & $0.109_{\pm 0.004}$ & $0.094_{\pm 0.001}$ & $0.096_{\pm 0.004}$ & \ul{$0.084_{\pm 0.004}$} \\
 & Pedestrian & $\textbf{0.071}_{\pm 0.012}$ & $0.096_{\pm 0.024}$ & $0.097_{\pm 0.006}$ & $0.086_{\pm 0.002}$ & $0.143_{\pm 0.007}$ & \ul{$0.079_{\pm 0.002}$} \\
 & Air & $\textbf{0.042}_{\pm 0.002}$ & $0.139_{\pm 0.025}$ & $0.171_{\pm 0.011}$ & $0.085_{\pm 0.001}$ & $0.092_{\pm 0.008}$ & \ul{$0.058_{\pm 0.006}$} \\
 & Temperature & $\textbf{0.142}_{\pm 0.010}$ & $0.189_{\pm 0.009}$ & $0.208_{\pm 0.012}$ & $0.259_{\pm 0.004}$ & $0.191_{\pm 0.004}$ & \ul{$0.156_{\pm 0.012}$} \\
 & Rain & $\textbf{0.067}_{\pm 0.015}$ & $0.228_{\pm 0.090}$ & $0.142_{\pm 0.020}$ & $0.228_{\pm 0.008}$ & $0.177_{\pm 0.016}$ & \ul{$0.104_{\pm 0.006}$} \\
 & NN5 & $\textbf{0.140}_{\pm 0.005}$ & $0.295_{\pm 0.021}$ & $0.240_{\pm 0.014}$ & $0.304_{\pm 0.010}$ & $0.220_{\pm 0.012}$ & \ul{$0.175_{\pm 0.011}$} \\
 & Fred-MD & $\textbf{0.021}_{\pm 0.002}$ & $0.079_{\pm 0.016}$ & $0.098_{\pm 0.024}$ & $0.104_{\pm 0.009}$ & $0.126_{\pm 0.009}$ & \ul{$0.076_{\pm 0.025}$} \\
 & Exchange & $0.358_{\pm 0.010}$ & \ul{$0.351_{\pm 0.098}$} & $\textbf{0.345}_{\pm 0.018}$ & $0.442_{\pm 0.020}$ & $0.515_{\pm 0.021}$ & $0.486_{\pm 0.040}$ \\ \bottomrule
\end{tabular}
\caption{Marginal Distribution Distance~(MDD) results of in-domain generation for sequence length 168. Best results are highlighted in bold face and second best results are underlined.}
\label{tab:indomain_mdd}
\end{table*}
}

%% file: src_figtabs/tab_app_ablamdd.tex
{\setlength{\tabcolsep}{4pt}
\begin{table*}[t]
\centering
\small
\begin{tabular}{@{}ccccccccc@{}}
\toprule
 &  & \multicolumn{5}{c}{TimeDP} & \multirow{2}{*}{- PAM} & \multirow{2}{*}{- Prompt} \\ \cmidrule(r){3-7}
 & \# Prototypes & 4 & 8 & 16 & 32 & 64 &  &  \\ \midrule
\multirow{13}{*}{\rotatebox{90}{Marginal Distribution Distance}} & Electricity & $0.006_{\pm 0.001}$ & $\textbf{0.005}_{\pm 0.001}$ & $0.005_{\pm 0.002}$ & $0.005_{\pm 0.002}$ & \ul{$0.005_{\pm 0.002}$} & $0.008_{\pm 0.004}$ & $0.014_{\pm 0.008}$ \\
 & Solar & $46.294_{\pm 19.573}$ & $45.134_{\pm 21.804}$ & $56.414_{\pm 21.890}$ & \ul{$38.374_{\pm 22.157}$} & $64.258_{\pm 29.679}$ & $47.291_{\pm 16.147}$ & $\textbf{32.032}_{\pm 6.336}$ \\
 & Wind & $0.089_{\pm 0.014}$ & $0.084_{\pm 0.011}$ & \ul{$0.084_{\pm 0.009}$} & $\textbf{0.084}_{\pm 0.012}$ & $0.084_{\pm 0.009}$ & $0.087_{\pm 0.006}$ & $0.090_{\pm 0.023}$ \\
 & Traffic & $0.072_{\pm 0.021}$ & $0.051_{\pm 0.003}$ & $\textbf{0.049}_{\pm 0.004}$ & \ul{$0.050_{\pm 0.002}$} & $0.050_{\pm 0.001}$ & $0.131_{\pm 0.001}$ & $0.133_{\pm 0.004}$ \\
 & Taxi & $0.087_{\pm 0.004}$ & $0.082_{\pm 0.005}$ & $\textbf{0.081}_{\pm 0.008}$ & $0.082_{\pm 0.008}$ & \ul{$0.081_{\pm 0.007}$} & $0.089_{\pm 0.002}$ & $0.089_{\pm 0.004}$ \\
 & Pedestrian & $0.080_{\pm 0.010}$ & $0.072_{\pm 0.006}$ & $\textbf{0.071}_{\pm 0.012}$ & $0.073_{\pm 0.009}$ & \ul{$0.072_{\pm 0.010}$} & $0.076_{\pm 0.003}$ & $0.076_{\pm 0.011}$ \\
 & Air & $0.050_{\pm 0.005}$ & $0.043_{\pm 0.003}$ & $\textbf{0.042}_{\pm 0.002}$ & $0.042_{\pm 0.003}$ & \ul{$0.042_{\pm 0.003}$} & $0.046_{\pm 0.002}$ & $0.048_{\pm 0.007}$ \\
 & Temperature & $0.153_{\pm 0.013}$ & \ul{$0.138_{\pm 0.007}$} & $0.142_{\pm 0.010}$ & $0.143_{\pm 0.011}$ & $0.144_{\pm 0.011}$ & $0.163_{\pm 0.002}$ & $\textbf{0.128}_{\pm 0.021}$ \\
 & Fred-MD & $0.024_{\pm 0.004}$ & $0.020_{\pm 0.003}$ & $0.021_{\pm 0.002}$ & \ul{$0.020_{\pm 0.002}$} & $\textbf{0.020}_{\pm 0.001}$ & $0.070_{\pm 0.002}$ & $0.029_{\pm 0.010}$ \\
 & Exchange & $0.358_{\pm 0.032}$ & $0.358_{\pm 0.008}$ & $0.358_{\pm 0.010}$ & \ul{$0.357_{\pm 0.006}$} & $0.360_{\pm 0.008}$ & $0.493_{\pm 0.003}$ & $\textbf{0.327}_{\pm 0.006}$ \\ \cmidrule(l){2-9} 
 & Average & 3.955 & 3.849 & 4.790 & \ul{3.286} & 5.443 & 4.060 & \textbf{2.769} \\ \bottomrule
\end{tabular}

\caption{Marginal Distribution Distance~(MDD) results for ablation study on generation sequence length 168. Best results are highlighted in bold face. Second best results are underlined.}
\label{tab:ablation_mdd}
\end{table*}
 }

%% file: src_figtabs/tab_app_id_24.tex
{\setlength{\tabcolsep}{4pt}
\begin{table*}[t]
\centering
\small
\begin{tabular}{@{}cccccccc@{}}
\toprule
 &  & \textbf{TimeDP} & \textbf{TimeGAN} & \textbf{GT-GAN} & \textbf{TimeVAE} & \textbf{TimeVQVAE} & \textbf{TimeVQVAE-C} \\ \midrule
\multirow{12}{*}{\rotatebox{90}{Maximum Mean Discrepancy}} & Electricity & $\textbf{0.001}_{\pm 0.001}$ & $0.270_{\pm 0.095}$ & $0.189_{\pm 0.085}$ & $0.980_{\pm 0.016}$ & $0.142_{\pm 0.017}$ & \ul{$0.002_{\pm 0.001}$} \\
 & Solar & $\textbf{0.023}_{\pm 0.016}$ & $0.513_{\pm 0.031}$ & $0.517_{\pm 0.027}$ & $0.321_{\pm 0.005}$ & $0.411_{\pm 0.004}$ & \ul{$0.036_{\pm 0.011}$} \\
 & Wind & \ul{$0.042_{\pm 0.019}$} & $0.348_{\pm 0.019}$ & $0.308_{\pm 0.015}$ & $0.419_{\pm 0.006}$ & $0.129_{\pm 0.008}$ & $\textbf{0.034}_{\pm 0.011}$ \\
 & Traffic & $\textbf{0.009}_{\pm 0.005}$ & $0.466_{\pm 0.071}$ & $0.522_{\pm 0.072}$ & $0.146_{\pm 0.005}$ & $0.189_{\pm 0.009}$ & \ul{$0.014_{\pm 0.004}$} \\
 & Taxi & $\textbf{0.032}_{\pm 0.005}$ & $0.156_{\pm 0.046}$ & $0.243_{\pm 0.030}$ & $0.049_{\pm 0.004}$ & \ul{$0.036_{\pm 0.003}$} & $0.036_{\pm 0.004}$ \\
 & Pedestrian & $\textbf{0.056}_{\pm 0.011}$ & $0.085_{\pm 0.022}$ & $0.121_{\pm 0.011}$ & $0.111_{\pm 0.003}$ & $0.087_{\pm 0.005}$ & \ul{$0.063_{\pm 0.004}$} \\
 & Air & $\textbf{0.008}_{\pm 0.004}$ & $0.146_{\pm 0.022}$ & $0.206_{\pm 0.007}$ & $0.192_{\pm 0.010}$ & \ul{$0.010_{\pm 0.002}$} & $0.017_{\pm 0.005}$ \\
 & Temperature & $\textbf{0.086}_{\pm 0.015}$ & $0.801_{\pm 0.048}$ & $0.801_{\pm 0.015}$ & $1.003_{\pm 0.012}$ & $0.215_{\pm 0.018}$ & \ul{$0.111_{\pm 0.019}$} \\
 & Rain & $\textbf{0.014}_{\pm 0.010}$ & $0.182_{\pm 0.096}$ & $0.085_{\pm 0.052}$ & $0.737_{\pm 0.048}$ & \ul{$0.036_{\pm 0.006}$} & $0.037_{\pm 0.004}$ \\
 & NN5 & $\textbf{0.072}_{\pm 0.015}$ & $0.704_{\pm 0.028}$ & $0.507_{\pm 0.078}$ & $0.708_{\pm 0.037}$ & $0.162_{\pm 0.011}$ & \ul{$0.086_{\pm 0.004}$} \\
 & Fred-MD & $\textbf{0.002}_{\pm 0.001}$ & $1.113_{\pm 0.048}$ & $0.822_{\pm 0.139}$ & $0.187_{\pm 0.014}$ & $0.003_{\pm 0.001}$ & \ul{$0.003_{\pm 0.000}$} \\
 & Exchange & $\textbf{0.127}_{\pm 0.066}$ & $0.504_{\pm 0.024}$ & $0.484_{\pm 0.074}$ & $0.662_{\pm 0.018}$ & $0.276_{\pm 0.026}$ & \ul{$0.155_{\pm 0.039}$} \\ \midrule
\multirow{12}{*}{\rotatebox{90}{K-L}} & Electricity & $\textbf{0.016}_{\pm 0.015}$ & $0.333_{\pm 0.134}$ & $0.423_{\pm 0.178}$ & $1.245_{\pm 0.018}$ & $0.302_{\pm 0.022}$ & \ul{$0.023_{\pm 0.007}$} \\
 & Solar & $\textbf{0.017}_{\pm 0.009}$ & $0.410_{\pm 0.042}$ & \ul{$0.102_{\pm 0.039}$} & $0.458_{\pm 0.016}$ & $0.878_{\pm 0.031}$ & $0.120_{\pm 0.036}$ \\
 & Wind & $\textbf{0.316}_{\pm 0.057}$ & $2.378_{\pm 0.989}$ & \ul{$0.317_{\pm 0.175}$} & $1.105_{\pm 0.019}$ & $0.566_{\pm 0.030}$ & $0.374_{\pm 0.053}$ \\
 & Traffic & \ul{$0.014_{\pm 0.005}$} & $1.061_{\pm 0.177}$ & $1.211_{\pm 0.286}$ & $0.446_{\pm 0.116}$ & $0.219_{\pm 0.016}$ & $\textbf{0.013}_{\pm 0.005}$ \\
 & Taxi & $\textbf{0.006}_{\pm 0.002}$ & $0.547_{\pm 0.150}$ & $0.872_{\pm 0.148}$ & $0.263_{\pm 0.022}$ & $0.121_{\pm 0.015}$ & \ul{$0.015_{\pm 0.007}$} \\
 & Pedestrian & $\textbf{0.016}_{\pm 0.010}$ & $0.369_{\pm 0.092}$ & $0.408_{\pm 0.105}$ & $0.211_{\pm 0.015}$ & $0.431_{\pm 0.017}$ & \ul{$0.023_{\pm 0.008}$} \\
 & Air & $\textbf{0.029}_{\pm 0.007}$ & $0.477_{\pm 0.154}$ & $0.923_{\pm 0.119}$ & $0.683_{\pm 0.067}$ & $0.108_{\pm 0.015}$ & \ul{$0.057_{\pm 0.006}$} \\
 & Temperature & $\textbf{0.240}_{\pm 0.029}$ & $7.672_{\pm 2.287}$ & $1.419_{\pm 0.461}$ & $1.288_{\pm 0.145}$ & $0.764_{\pm 0.076}$ & \ul{$0.357_{\pm 0.055}$} \\
 & Rain & $\textbf{0.037}_{\pm 0.019}$ & $0.356_{\pm 0.089}$ & $0.468_{\pm 0.158}$ & $0.987_{\pm 0.073}$ & \ul{$0.077_{\pm 0.020}$} & $0.089_{\pm 0.037}$ \\
 & NN5 & \ul{$0.074_{\pm 0.028}$} & $2.273_{\pm 0.231}$ & $1.154_{\pm 0.288}$ & $1.434_{\pm 0.117}$ & $0.904_{\pm 0.175}$ & $\textbf{0.063}_{\pm 0.010}$ \\
 & Fred-MD & $\textbf{0.515}_{\pm 0.174}$ & $3.356_{\pm 0.639}$ & $2.161_{\pm 0.483}$ & $0.701_{\pm 0.091}$ & \ul{$0.533_{\pm 0.067}$} & $0.574_{\pm 0.168}$ \\
 & Exchange & $7.270_{\pm 2.080}$ & $5.482_{\pm 2.108}$ & $\textbf{3.897}_{\pm 2.172}$ & $8.682_{\pm 1.334}$ & \ul{$4.201_{\pm 0.945}$} & $4.561_{\pm 0.212}$ \\ \midrule
\multirow{12}{*}{\rotatebox{90}{Marginal Distribution Distance}} & Electricity & \ul{$0.004_{\pm 0.002}$} & $0.059_{\pm 0.011}$ & $0.052_{\pm 0.014}$ & $0.114_{\pm 0.001}$ & $0.063_{\pm 0.002}$ & $\textbf{0.004}_{\pm 0.001}$ \\
 & Solar & $70.117_{\pm 16.651}$ & $64.855_{\pm 3.985}$ & $76.076_{\pm 4.681}$ & $\textbf{46.304}_{\pm 0.010}$ & \ul{$50.554_{\pm 0.146}$} & $53.143_{\pm 0.574}$ \\
 & Wind & $\textbf{0.115}_{\pm 0.017}$ & $0.195_{\pm 0.016}$ & $0.138_{\pm 0.013}$ & $0.291_{\pm 0.003}$ & $0.188_{\pm 0.007}$ & \ul{$0.133_{\pm 0.010}$} \\
 & Traffic & \ul{$0.017_{\pm 0.005}$} & $0.112_{\pm 0.008}$ & $0.124_{\pm 0.003}$ & $0.079_{\pm 0.002}$ & $0.080_{\pm 0.002}$ & $\textbf{0.016}_{\pm 0.002}$ \\
 & Taxi & \ul{$0.016_{\pm 0.002}$} & $0.036_{\pm 0.005}$ & $0.049_{\pm 0.002}$ & $0.028_{\pm 0.001}$ & $0.029_{\pm 0.002}$ & $\textbf{0.014}_{\pm 0.001}$ \\
 & Pedestrian & \ul{$0.049_{\pm 0.006}$} & $0.060_{\pm 0.004}$ & $0.075_{\pm 0.002}$ & $0.069_{\pm 0.002}$ & $0.092_{\pm 0.002}$ & $\textbf{0.046}_{\pm 0.001}$ \\
 & Air & $\textbf{0.028}_{\pm 0.005}$ & $0.076_{\pm 0.016}$ & $0.112_{\pm 0.003}$ & $0.100_{\pm 0.003}$ & $0.053_{\pm 0.003}$ & \ul{$0.035_{\pm 0.003}$} \\
 & Temperature & $\textbf{0.102}_{\pm 0.007}$ & $0.145_{\pm 0.004}$ & $0.175_{\pm 0.009}$ & $0.180_{\pm 0.002}$ & $0.157_{\pm 0.006}$ & \ul{$0.109_{\pm 0.007}$} \\
 & Rain & $\textbf{0.007}_{\pm 0.001}$ & $0.063_{\pm 0.017}$ & $0.046_{\pm 0.011}$ & $0.122_{\pm 0.002}$ & $0.052_{\pm 0.001}$ & \ul{$0.012_{\pm 0.003}$} \\
 & NN5 & \ul{$0.073_{\pm 0.005}$} & $0.248_{\pm 0.005}$ & $0.185_{\pm 0.019}$ & $0.243_{\pm 0.008}$ & $0.111_{\pm 0.004}$ & $\textbf{0.068}_{\pm 0.003}$ \\
 & Fred-MD & $\textbf{0.010}_{\pm 0.002}$ & $0.107_{\pm 0.002}$ & $0.098_{\pm 0.004}$ & $0.053_{\pm 0.007}$ & $0.038_{\pm 0.003}$ & \ul{$0.011_{\pm 0.001}$} \\
 & Exchange & $0.348_{\pm 0.011}$ & \ul{$0.321_{\pm 0.040}$} & $\textbf{0.313}_{\pm 0.017}$ & $0.417_{\pm 0.009}$ & $0.466_{\pm 0.009}$ & $0.441_{\pm 0.021}$ \\ \bottomrule
\end{tabular}
\caption{Maximum mean discrepancy~(MMD) and K-L divergence~(K-L) and Marginal Distribution Distance~(MDD) results of in-domain generation for sequence length 24. Best results are highlighted in bold face and second best results are underlined.}
\label{tab:indomain_24}
\end{table*}
}

%% file: src_figtabs/tab_app_id_96.tex
{\setlength{\tabcolsep}{4pt}
\begin{table*}[t]
\centering
\small
\begin{tabular}{@{}cccccccc@{}}
\toprule
 &  & \textbf{TimeDP} & \textbf{TimeGAN} & \textbf{GT-GAN} & \textbf{TimeVAE} & \textbf{TimeVQVAE} & \textbf{TimeVQVAE-C} \\ \midrule
\multirow{12}{*}{\rotatebox{90}{Maximum Mean Discrepancy}} & Electricity & $\textbf{0.002}_{\pm   0.001}$ & $0.401_{\pm 0.219}$ & $0.317_{\pm 0.311}$ & $0.648_{\pm 0.012}$ & $0.153_{\pm 0.023}$ & \ul{$0.002_{\pm 0.000}$} \\
 & Solar & $\textbf{0.019}_{\pm 0.004}$ & $0.584_{\pm 0.056}$ & $0.634_{\pm 0.119}$ & $0.369_{\pm 0.012}$ & $0.439_{\pm 0.021}$ & \ul{$0.035_{\pm 0.008}$} \\
 & Wind & \ul{$0.029_{\pm 0.015}$} & $0.195_{\pm 0.030}$ & $0.217_{\pm 0.134}$ & $0.179_{\pm 0.002}$ & $0.129_{\pm 0.016}$ & $\textbf{0.013}_{\pm 0.007}$ \\
 & Traffic & $\textbf{0.018}_{\pm 0.005}$ & $0.456_{\pm 0.100}$ & $0.513_{\pm 0.073}$ & $0.162_{\pm 0.005}$ & $0.189_{\pm 0.014}$ & \ul{$0.027_{\pm 0.002}$} \\
 & Taxi & $\textbf{0.055}_{\pm 0.007}$ & $0.197_{\pm 0.057}$ & $0.328_{\pm 0.071}$ & \ul{$0.060_{\pm 0.002}$} & $0.089_{\pm 0.005}$ & $0.061_{\pm 0.002}$ \\
 & Pedestrian & $\textbf{0.040}_{\pm 0.010}$ & $0.068_{\pm 0.025}$ & $0.159_{\pm 0.149}$ & $0.064_{\pm 0.001}$ & $0.066_{\pm 0.011}$ & \ul{$0.049_{\pm 0.005}$} \\
 & Air & $\textbf{0.010}_{\pm 0.006}$ & $0.103_{\pm 0.046}$ & $0.244_{\pm 0.160}$ & $0.085_{\pm 0.011}$ & \ul{$0.022_{\pm 0.003}$} & $0.036_{\pm 0.007}$ \\
 & Temperature & $\textbf{0.178}_{\pm 0.047}$ & $0.827_{\pm 0.092}$ & $0.859_{\pm 0.109}$ & $0.958_{\pm 0.012}$ & $0.413_{\pm 0.034}$ & \ul{$0.205_{\pm 0.033}$} \\
 & Rain & $\textbf{0.044}_{\pm 0.027}$ & $0.285_{\pm 0.199}$ & $0.166_{\pm 0.203}$ & $0.327_{\pm 0.020}$ & \ul{$0.068_{\pm 0.008}$} & $0.072_{\pm 0.001}$ \\
 & NN5 & $\textbf{0.234}_{\pm 0.047}$ & $0.888_{\pm 0.091}$ & $0.706_{\pm 0.116}$ & $0.875_{\pm 0.053}$ & $0.321_{\pm 0.008}$ & \ul{$0.241_{\pm 0.011}$} \\
 & Fred-MD & \ul{$0.005_{\pm 0.002}$} & $0.055_{\pm 0.022}$ & $0.150_{\pm 0.184}$ & $0.046_{\pm 0.013}$ & $\textbf{0.004}_{\pm 0.001}$ & $0.006_{\pm 0.001}$ \\
 & Exchange & $0.343_{\pm 0.210}$ & $0.449_{\pm 0.018}$ & $0.567_{\pm 0.141}$ & $0.592_{\pm 0.124}$ & \ul{$0.337_{\pm 0.056}$} & $\textbf{0.224}_{\pm 0.028}$ \\ \midrule
\multirow{12}{*}{\rotatebox{90}{K-L}} & Electricity & $\textbf{0.015}_{\pm 0.017}$ & $0.552_{\pm 0.241}$ & $0.447_{\pm 0.147}$ & $0.790_{\pm 0.018}$ & $0.245_{\pm 0.039}$ & \ul{$0.031_{\pm 0.005}$} \\
 & Solar & $\textbf{0.010}_{\pm 0.002}$ & $0.922_{\pm 0.953}$ & \ul{$0.117_{\pm 0.047}$} & $0.288_{\pm 0.015}$ & $0.812_{\pm 0.077}$ & $0.151_{\pm 0.114}$ \\
 & Wind & \ul{$0.182_{\pm 0.029}$} & $2.288_{\pm 1.135}$ & $\textbf{0.164}_{\pm 0.071}$ & $0.527_{\pm 0.009}$ & $0.510_{\pm 0.112}$ & $0.284_{\pm 0.115}$ \\
 & Traffic & $\textbf{0.010}_{\pm 0.006}$ & $1.241_{\pm 0.757}$ & $1.198_{\pm 0.340}$ & $0.205_{\pm 0.008}$ & $0.192_{\pm 0.016}$ & \ul{$0.014_{\pm 0.004}$} \\
 & Taxi & $\textbf{0.010}_{\pm 0.007}$ & $0.550_{\pm 0.333}$ & $0.801_{\pm 0.270}$ & $0.121_{\pm 0.003}$ & $0.102_{\pm 0.017}$ & \ul{$0.027_{\pm 0.007}$} \\
 & Pedestrian & $\textbf{0.012}_{\pm 0.005}$ & $0.458_{\pm 0.368}$ & $0.327_{\pm 0.148}$ & $0.074_{\pm 0.004}$ & $0.373_{\pm 0.027}$ & \ul{$0.034_{\pm 0.006}$} \\
 & Air & $\textbf{0.025}_{\pm 0.011}$ & $0.317_{\pm 0.102}$ & $0.510_{\pm 0.147}$ & $0.168_{\pm 0.012}$ & \ul{$0.061_{\pm 0.012}$} & $0.096_{\pm 0.019}$ \\
 & Temperature & $\textbf{0.392}_{\pm 0.097}$ & $11.298_{\pm 1.212}$ & $5.830_{\pm 4.888}$ & $1.737_{\pm 0.068}$ & $1.135_{\pm 0.110}$ & \ul{$0.567_{\pm 0.050}$} \\
 & Rain & $\textbf{0.022}_{\pm 0.009}$ & $0.549_{\pm 0.257}$ & $0.481_{\pm 0.146}$ & $0.237_{\pm 0.025}$ & \ul{$0.055_{\pm 0.021}$} & $0.091_{\pm 0.010}$ \\
 & NN5 & $\textbf{0.069}_{\pm 0.033}$ & $5.829_{\pm 6.120}$ & $2.709_{\pm 2.080}$ & $1.537_{\pm 0.224}$ & $1.038_{\pm 0.168}$ & \ul{$0.103_{\pm 0.072}$} \\
 & Fred-MD & $1.183_{\pm 0.280}$ & $0.628_{\pm 0.335}$ & \ul{$0.486_{\pm 0.108}$} & $\textbf{0.441}_{\pm 0.063}$ & $0.813_{\pm 0.093}$ & $1.240_{\pm 0.232}$ \\
 & Exchange & $14.336_{\pm 2.652}$ & $6.445_{\pm 2.210}$ & $10.170_{\pm 6.381}$ & $9.937_{\pm 4.322}$ & $\textbf{3.740}_{\pm 1.261}$ & \ul{$4.408_{\pm 0.700}$} \\ \midrule
\multirow{12}{*}{\rotatebox{90}{Marginal Distribution Distance}} & Electricity & \ul{$0.005_{\pm 0.002}$} & $0.070_{\pm 0.020}$ & $0.046_{\pm 0.011}$ & $0.082_{\pm 0.002}$ & $0.058_{\pm 0.002}$ & $\textbf{0.005}_{\pm 0.001}$ \\
 & Solar & $77.545_{\pm 18.384}$ & $60.819_{\pm 9.001}$ & $77.198_{\pm 5.522}$ & $\textbf{47.123}_{\pm 0.114}$ & \ul{$50.619_{\pm 0.099}$} & $53.220_{\pm 0.832}$ \\
 & Wind & \ul{$0.089_{\pm 0.012}$} & $0.207_{\pm 0.054}$ & $0.135_{\pm 0.035}$ & $0.208_{\pm 0.002}$ & $0.159_{\pm 0.009}$ & $\textbf{0.069}_{\pm 0.014}$ \\
 & Traffic & $\textbf{0.024}_{\pm 0.003}$ & $0.109_{\pm 0.018}$ & $0.122_{\pm 0.009}$ & $0.083_{\pm 0.001}$ & $0.089_{\pm 0.002}$ & \ul{$0.028_{\pm 0.002}$} \\
 & Taxi & \ul{$0.043_{\pm 0.004}$} & $0.072_{\pm 0.008}$ & $0.092_{\pm 0.010}$ & $0.064_{\pm 0.001}$ & $0.070_{\pm 0.003}$ & $\textbf{0.040}_{\pm 0.002}$ \\
 & Pedestrian & $\textbf{0.058}_{\pm 0.007}$ & $0.074_{\pm 0.018}$ & $0.088_{\pm 0.025}$ & $0.073_{\pm 0.001}$ & $0.116_{\pm 0.003}$ & \ul{$0.058_{\pm 0.003}$} \\
 & Air & $\textbf{0.031}_{\pm 0.007}$ & $0.091_{\pm 0.021}$ & $0.145_{\pm 0.022}$ & $0.070_{\pm 0.002}$ & $0.066_{\pm 0.004}$ & \ul{$0.056_{\pm 0.007}$} \\
 & Temperature & $\textbf{0.135}_{\pm 0.016}$ & $0.154_{\pm 0.004}$ & $0.185_{\pm 0.007}$ & $0.196_{\pm 0.001}$ & $0.198_{\pm 0.006}$ & \ul{$0.143_{\pm 0.010}$} \\
 & Rain & $\textbf{0.039}_{\pm 0.006}$ & $0.192_{\pm 0.059}$ & $0.113_{\pm 0.011}$ & $0.196_{\pm 0.003}$ & $0.134_{\pm 0.009}$ & \ul{$0.047_{\pm 0.012}$} \\
 & NN5 & \ul{$0.175_{\pm 0.016}$} & $0.270_{\pm 0.014}$ & $0.225_{\pm 0.017}$ & $0.268_{\pm 0.006}$ & $0.196_{\pm 0.007}$ & $\textbf{0.158}_{\pm 0.004}$ \\
 & Fred-MD & $\textbf{0.023}_{\pm 0.005}$ & $0.063_{\pm 0.009}$ & $0.076_{\pm 0.023}$ & $0.055_{\pm 0.012}$ & $0.083_{\pm 0.004}$ & \ul{$0.045_{\pm 0.017}$} \\
 & Exchange & $0.402_{\pm 0.083}$ & \ul{$0.342_{\pm 0.044}$} & $\textbf{0.334}_{\pm 0.024}$ & $0.425_{\pm 0.027}$ & $0.503_{\pm 0.016}$ & $0.426_{\pm 0.033}$ \\ \bottomrule
\end{tabular}
\caption{Maximum mean discrepancy~(MMD) and K-L divergence~(K-L) and Marginal Distribution Distance~(MDD) results of in-domain generation for sequence length 96. Best results are highlighted in bold face and second best results are underlined.}
\label{tab:indomain_96}
\end{table*}
}

%% file: src_figtabs/tab_app_id_336.tex
{\setlength{\tabcolsep}{4pt}
\begin{table*}[t]
\centering
\small
\begin{tabular}{@{}cccccccc@{}}
\toprule
 &  & \textbf{TimeDP} & \textbf{TimeGAN} & \textbf{GT-GAN} & \textbf{TimeVAE} & \textbf{TimeVQVAE} & \textbf{TimeVQVAE-C} \\ \midrule
\multirow{12}{*}{\rotatebox{90}{Maximum Mean Discrepancy}} & Electricity & $\textbf{0.001}_{\pm   0.002}$ & $0.320_{\pm 0.197}$ & $0.347_{\pm 0.524}$ & $0.504_{\pm 0.012}$ & $0.135_{\pm 0.018}$ & \ul{$0.003_{\pm 0.000}$} \\
 & Solar & $\textbf{0.074}_{\pm 0.012}$ & $0.646_{\pm 0.032}$ & $0.735_{\pm 0.045}$ & $0.400_{\pm 0.015}$ & $0.493_{\pm 0.024}$ & \ul{$0.081_{\pm 0.006}$} \\
 & Wind & \ul{$0.024_{\pm 0.003}$} & $0.208_{\pm 0.095}$ & $0.221_{\pm 0.010}$ & $0.145_{\pm 0.005}$ & $0.139_{\pm 0.015}$ & $\textbf{0.014}_{\pm 0.007}$ \\
 & Traffic & $\textbf{0.077}_{\pm 0.009}$ & $0.518_{\pm 0.028}$ & $0.704_{\pm 0.101}$ & $0.235_{\pm 0.011}$ & $0.233_{\pm 0.017}$ & \ul{$0.091_{\pm 0.005}$} \\
 & Taxi & $0.189_{\pm 0.034}$ & $0.329_{\pm 0.011}$ & $0.390_{\pm 0.078}$ & $\textbf{0.166}_{\pm 0.002}$ & $0.202_{\pm 0.004}$ & \ul{$0.189_{\pm 0.007}$} \\
 & Pedestrian & $\textbf{0.035}_{\pm 0.008}$ & $0.140_{\pm 0.067}$ & $0.158_{\pm 0.019}$ & $0.061_{\pm 0.003}$ & $0.074_{\pm 0.007}$ & \ul{$0.067_{\pm 0.008}$} \\
 & Air & $\textbf{0.018}_{\pm 0.005}$ & $0.213_{\pm 0.037}$ & $0.261_{\pm 0.062}$ & $0.075_{\pm 0.005}$ & \ul{$0.038_{\pm 0.003}$} & $0.053_{\pm 0.007}$ \\
 & Temperature & $\textbf{0.185}_{\pm 0.021}$ & $0.914_{\pm 0.031}$ & $0.934_{\pm 0.051}$ & $0.976_{\pm 0.022}$ & $0.318_{\pm 0.008}$ & \ul{$0.197_{\pm 0.014}$} \\
 & Rain & $0.096_{\pm 0.086}$ & $0.151_{\pm 0.108}$ & $0.198_{\pm 0.295}$ & $0.135_{\pm 0.018}$ & $\textbf{0.066}_{\pm 0.002}$ & \ul{$0.081_{\pm 0.005}$} \\
 & NN5 & $\textbf{0.191}_{\pm 0.016}$ & $0.663_{\pm 0.027}$ & $0.951_{\pm 0.069}$ & $0.806_{\pm 0.086}$ & $0.333_{\pm 0.009}$ & \ul{$0.260_{\pm 0.049}$} \\
 & Fred-MD & $\textbf{0.006}_{\pm 0.004}$ & $0.156_{\pm 0.097}$ & $0.063_{\pm 0.006}$ & $0.052_{\pm 0.020}$ & $0.010_{\pm 0.006}$ & \ul{$0.007_{\pm 0.001}$} \\
 & Exchange & $0.683_{\pm 0.217}$ & $0.724_{\pm 0.130}$ & $0.553_{\pm 0.238}$ & $0.598_{\pm 0.166}$ & \ul{$0.318_{\pm 0.077}$} & $\textbf{0.275}_{\pm 0.198}$ \\ \midrule
\multirow{12}{*}{\rotatebox{90}{K-L}} & Electricity & $\textbf{0.016}_{\pm 0.016}$ & $0.396_{\pm 0.108}$ & $1.169_{\pm 2.029}$ & $0.719_{\pm 0.020}$ & $0.257_{\pm 0.023}$ & \ul{$0.042_{\pm 0.033}$} \\
 & Solar & $\textbf{0.017}_{\pm 0.006}$ & \ul{$0.179_{\pm 0.062}$} & $2.150_{\pm 2.179}$ & $0.249_{\pm 0.012}$ & $0.870_{\pm 0.110}$ & $0.201_{\pm 0.017}$ \\
 & Wind & \ul{$0.158_{\pm 0.032}$} & $0.169_{\pm 0.064}$ & $4.264_{\pm 2.790}$ & $0.373_{\pm 0.019}$ & $0.495_{\pm 0.043}$ & $\textbf{0.121}_{\pm 0.037}$ \\
 & Traffic & $\textbf{0.008}_{\pm 0.004}$ & $1.443_{\pm 0.354}$ & $2.992_{\pm 0.592}$ & $0.241_{\pm 0.012}$ & $0.210_{\pm 0.034}$ & \ul{$0.023_{\pm 0.006}$} \\
 & Taxi & $\textbf{0.076}_{\pm 0.034}$ & $0.937_{\pm 0.151}$ & $1.461_{\pm 0.367}$ & $0.160_{\pm 0.015}$ & $0.230_{\pm 0.026}$ & \ul{$0.094_{\pm 0.007}$} \\
 & Pedestrian & $\textbf{0.009}_{\pm 0.003}$ & $0.452_{\pm 0.131}$ & $1.544_{\pm 0.908}$ & \ul{$0.044_{\pm 0.007}$} & $0.407_{\pm 0.069}$ & $0.053_{\pm 0.017}$ \\
 & Air & $\textbf{0.017}_{\pm 0.010}$ & $1.031_{\pm 0.145}$ & $1.481_{\pm 0.553}$ & $0.180_{\pm 0.015}$ & $0.121_{\pm 0.028}$ & \ul{$0.079_{\pm 0.016}$} \\
 & Temperature & $\textbf{0.037}_{\pm 0.016}$ & $2.479_{\pm 1.867}$ & $9.139_{\pm 2.247}$ & $1.726_{\pm 0.122}$ & $1.119_{\pm 0.179}$ & \ul{$0.224_{\pm 0.123}$} \\
 & Rain & $\textbf{0.011}_{\pm 0.002}$ & $0.388_{\pm 0.112}$ & $0.271_{\pm 0.046}$ & $0.028_{\pm 0.005}$ & \ul{$0.025_{\pm 0.013}$} & $0.028_{\pm 0.005}$ \\
 & NN5 & $\textbf{0.075}_{\pm 0.034}$ & $1.418_{\pm 0.122}$ & $8.449_{\pm 5.473}$ & $1.233_{\pm 0.256}$ & $0.836_{\pm 0.250}$ & \ul{$0.180_{\pm 0.181}$} \\
 & Fred-MD & $0.815_{\pm 0.630}$ & \ul{$0.309_{\pm 0.145}$} & $0.568_{\pm 0.321}$ & $\textbf{0.212}_{\pm 0.047}$ & $0.586_{\pm 0.161}$ & $1.099_{\pm 0.020}$ \\
 & Exchange & $18.426_{\pm 4.493}$ & $14.116_{\pm 3.798}$ & $13.875_{\pm 2.965}$ & $10.670_{\pm 5.335}$ & \ul{$7.755_{\pm 2.566}$} & $\textbf{7.651}_{\pm 3.841}$ \\ \midrule
\multirow{12}{*}{\rotatebox{90}{Marginal Distribution Distance}} & Electricity & $\textbf{0.006}_{\pm 0.004}$ & $0.048_{\pm 0.009}$ & $0.061_{\pm 0.058}$ & $0.099_{\pm 0.002}$ & $0.069_{\pm 0.003}$ & \ul{$0.008_{\pm 0.003}$} \\
 & Solar & \ul{$49.170_{\pm 21.121}$} & $82.860_{\pm 4.163}$ & $74.403_{\pm 18.119}$ & $\textbf{13.755}_{\pm 0.077}$ & $54.065_{\pm 0.183}$ & $55.878_{\pm 0.249}$ \\
 & Wind & \ul{$0.091_{\pm 0.007}$} & $0.146_{\pm 0.028}$ & $0.251_{\pm 0.060}$ & $0.172_{\pm 0.004}$ & $0.168_{\pm 0.009}$ & $\textbf{0.070}_{\pm 0.009}$ \\
 & Traffic & $\textbf{0.066}_{\pm 0.002}$ & $0.191_{\pm 0.005}$ & $0.206_{\pm 0.031}$ & $0.134_{\pm 0.001}$ & $0.158_{\pm 0.005}$ & \ul{$0.072_{\pm 0.005}$} \\
 & Taxi & $\textbf{0.680}_{\pm 0.021}$ & $1.073_{\pm 0.162}$ & $0.986_{\pm 0.126}$ & $0.764_{\pm 0.003}$ & $0.750_{\pm 0.003}$ & \ul{$0.707_{\pm 0.001}$} \\
 & Pedestrian & $\textbf{0.074}_{\pm 0.006}$ & $0.121_{\pm 0.027}$ & $0.153_{\pm 0.070}$ & \ul{$0.086_{\pm 0.002}$} & $0.161_{\pm 0.007}$ & $0.098_{\pm 0.006}$ \\
 & Air & $\textbf{0.049}_{\pm 0.004}$ & $0.157_{\pm 0.002}$ & $0.166_{\pm 0.039}$ & $0.078_{\pm 0.002}$ & $0.099_{\pm 0.005}$ & \ul{$0.056_{\pm 0.005}$} \\
 & Temperature & $\textbf{0.153}_{\pm 0.010}$ & $0.230_{\pm 0.007}$ & $0.218_{\pm 0.013}$ & $0.341_{\pm 0.005}$ & $0.204_{\pm 0.009}$ & \ul{$0.155_{\pm 0.006}$} \\
 & Rain & $\textbf{0.100}_{\pm 0.007}$ & $0.144_{\pm 0.011}$ & $0.238_{\pm 0.184}$ & $0.210_{\pm 0.014}$ & $0.222_{\pm 0.009}$ & \ul{$0.138_{\pm 0.014}$} \\
 & NN5 & \ul{$1.079_{\pm 0.016}$} & $1.128_{\pm 0.017}$ & $1.199_{\pm 0.024}$ & $1.209_{\pm 0.021}$ & $1.109_{\pm 0.006}$ & $\textbf{1.074}_{\pm 0.012}$ \\
 & Fred-MD & $\textbf{0.031}_{\pm 0.007}$ & $0.106_{\pm 0.038}$ & $0.085_{\pm 0.006}$ & $0.093_{\pm 0.035}$ & $0.128_{\pm 0.016}$ & \ul{$0.060_{\pm 0.025}$} \\
 & Exchange & $0.510_{\pm 0.131}$ & $\textbf{0.337}_{\pm 0.023}$ & \ul{$0.383_{\pm 0.091}$} & $0.443_{\pm 0.027}$ & $0.537_{\pm 0.036}$ & $0.452_{\pm 0.048}$ \\ \bottomrule
\end{tabular}
\caption{Maximum mean discrepancy~(MMD) and K-L divergence~(K-L) and Marginal Distribution Distance~(MDD) results of in-domain generation for sequence length 336. Best results are highlighted in bold face and second best results are underlined.}
\label{tab:indomain_336}
\end{table*}
}

%% file: src_figtabs/fig_prompt_sem.tex
\begin{figure*}[!t]
  \centering
  \includegraphics[width=0.95\textwidth]{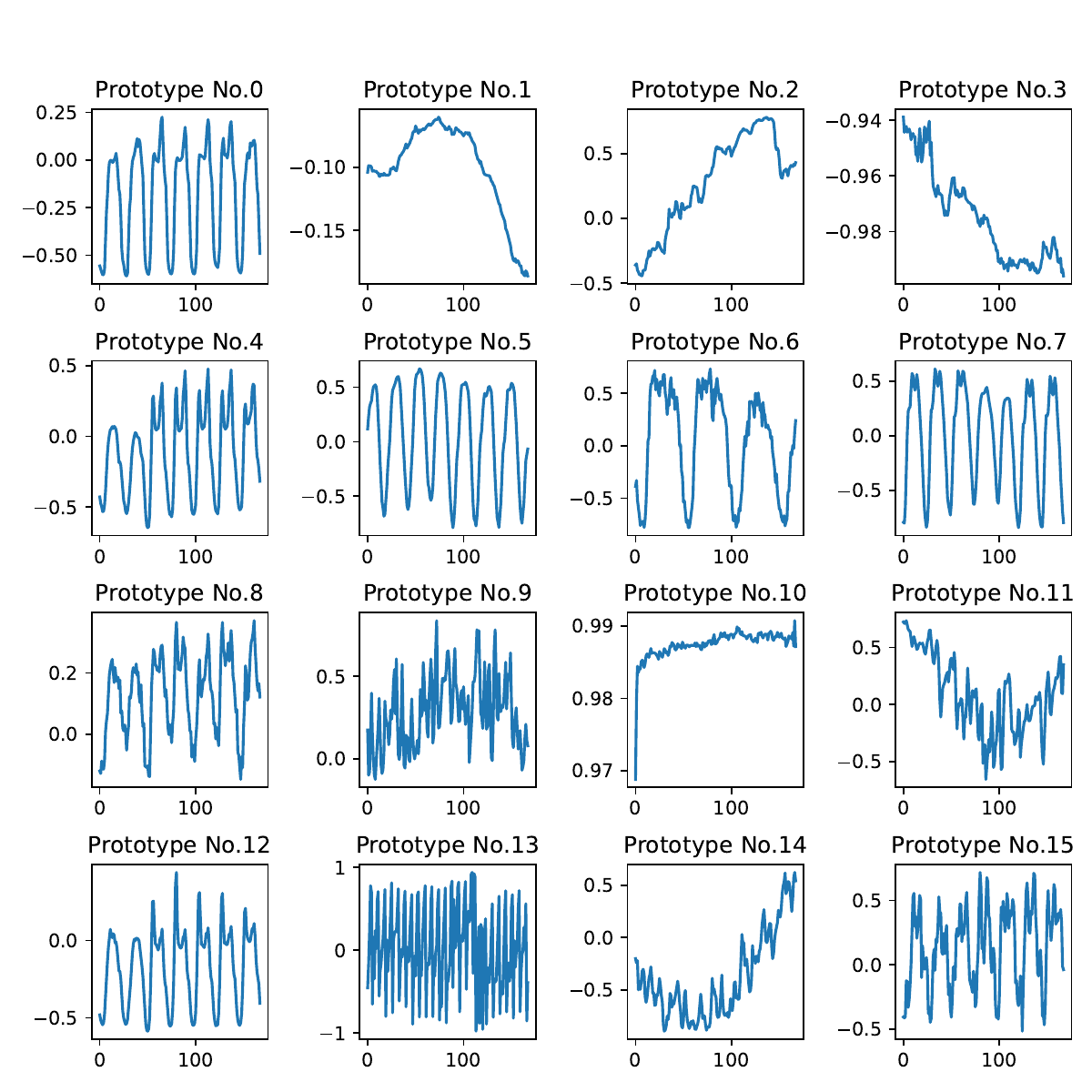}
  \caption{Semantic visualization of time series prototypes.}
  \label{fig:dp_semantic}
\end{figure*}

%% file: src_figtabs/fig_prompt_heatmap.tex
\begin{figure*}[!t]
  \centering
  \includegraphics[width=0.95\textwidth]{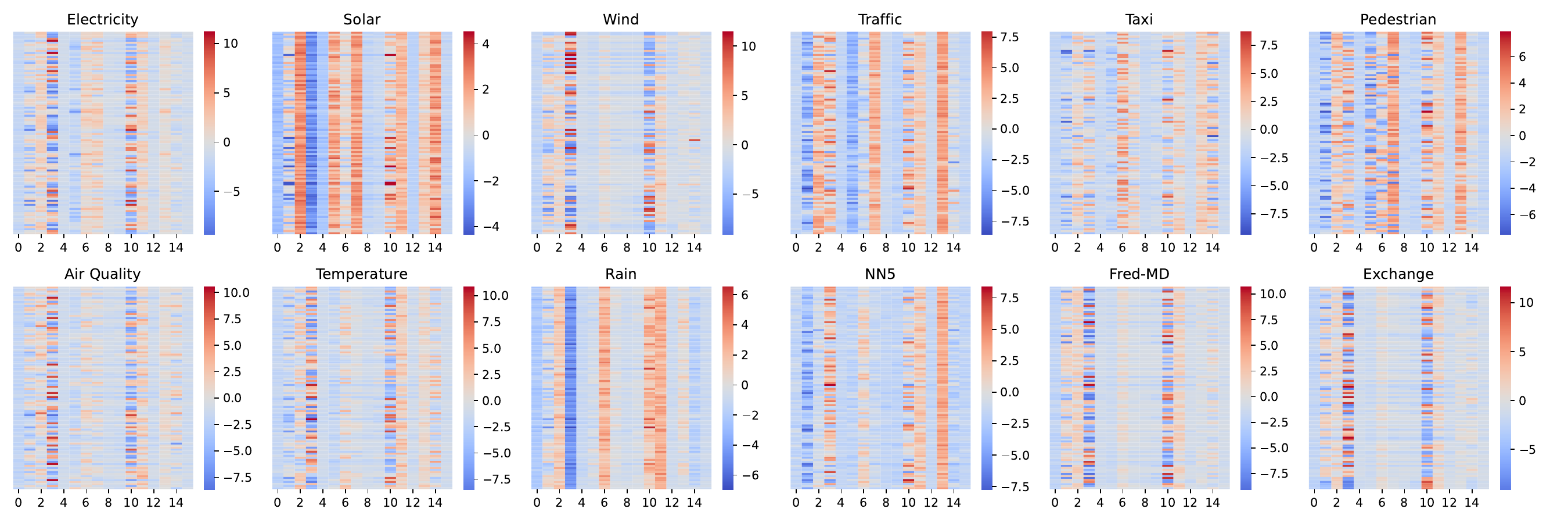}
  \caption{Heatmap of Domain Prompts.}
  \label{fig:dp_heat}
\end{figure*}

%% file: src_figtabs/fig_data_tsne.tex
\begin{figure*}[!t]
  \centering
  \includegraphics[width=0.55\textwidth]{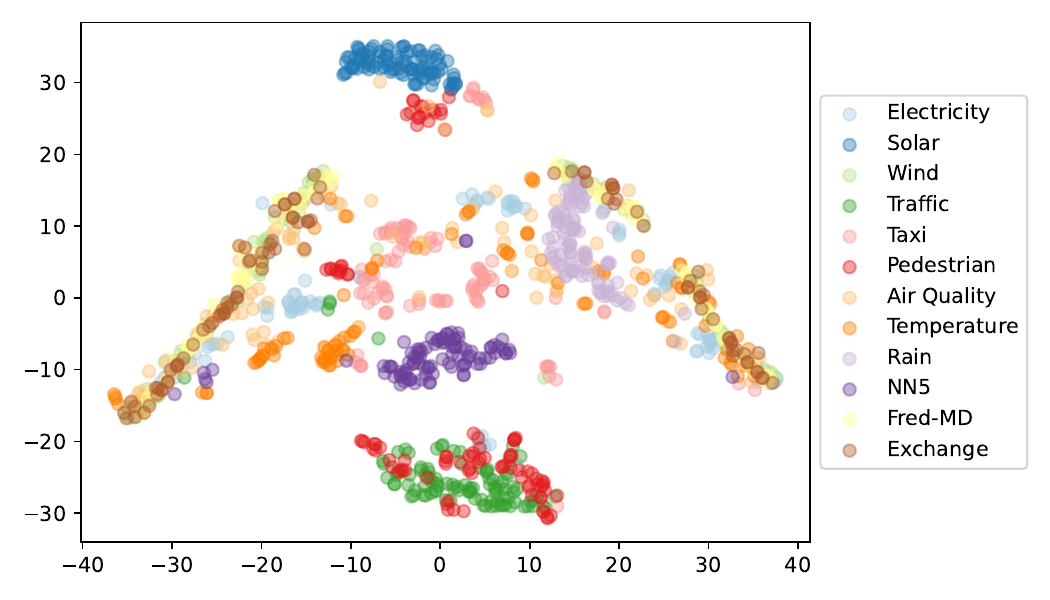}
  \caption{T-SNE visualization of domain prompts. Domain prompt generated for each dataset are marked with the same color.}
  \label{fig:dp_dataset_heat}
\end{figure*}

%% file: src_figtabs/fig_domain_tsne.tex
\begin{figure*}[!t]
  \centering
  \includegraphics[width=0.45\textwidth]{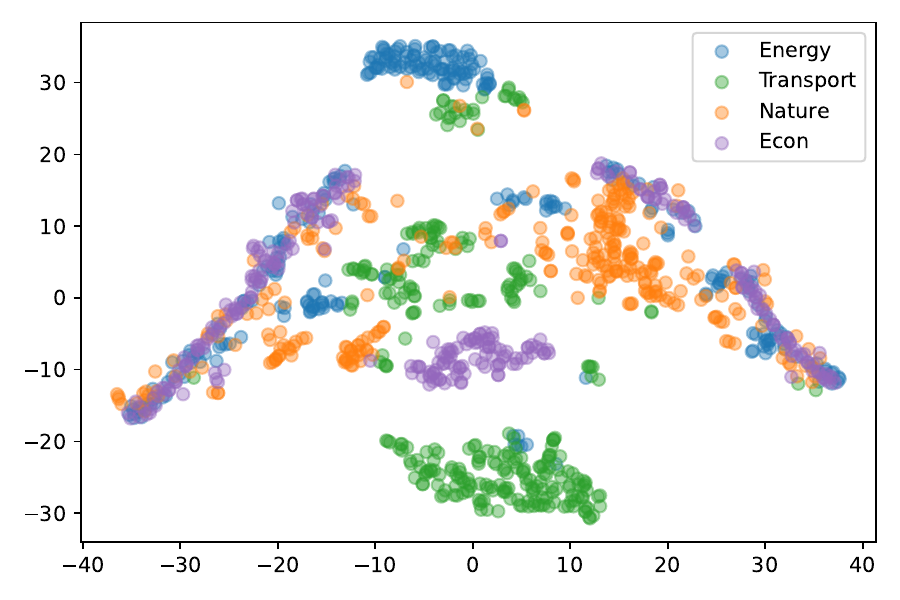}
  \caption{T-SNE visualization of domain prompts. Datasets in the same domain are marked with the same color.}
  \label{fig:dp_domain_heat}
\end{figure*}

%% file: src_figtabs/fig_web_zs.tex
\begin{figure*}[!t]
  \centering
  \includegraphics[width=0.95\textwidth]{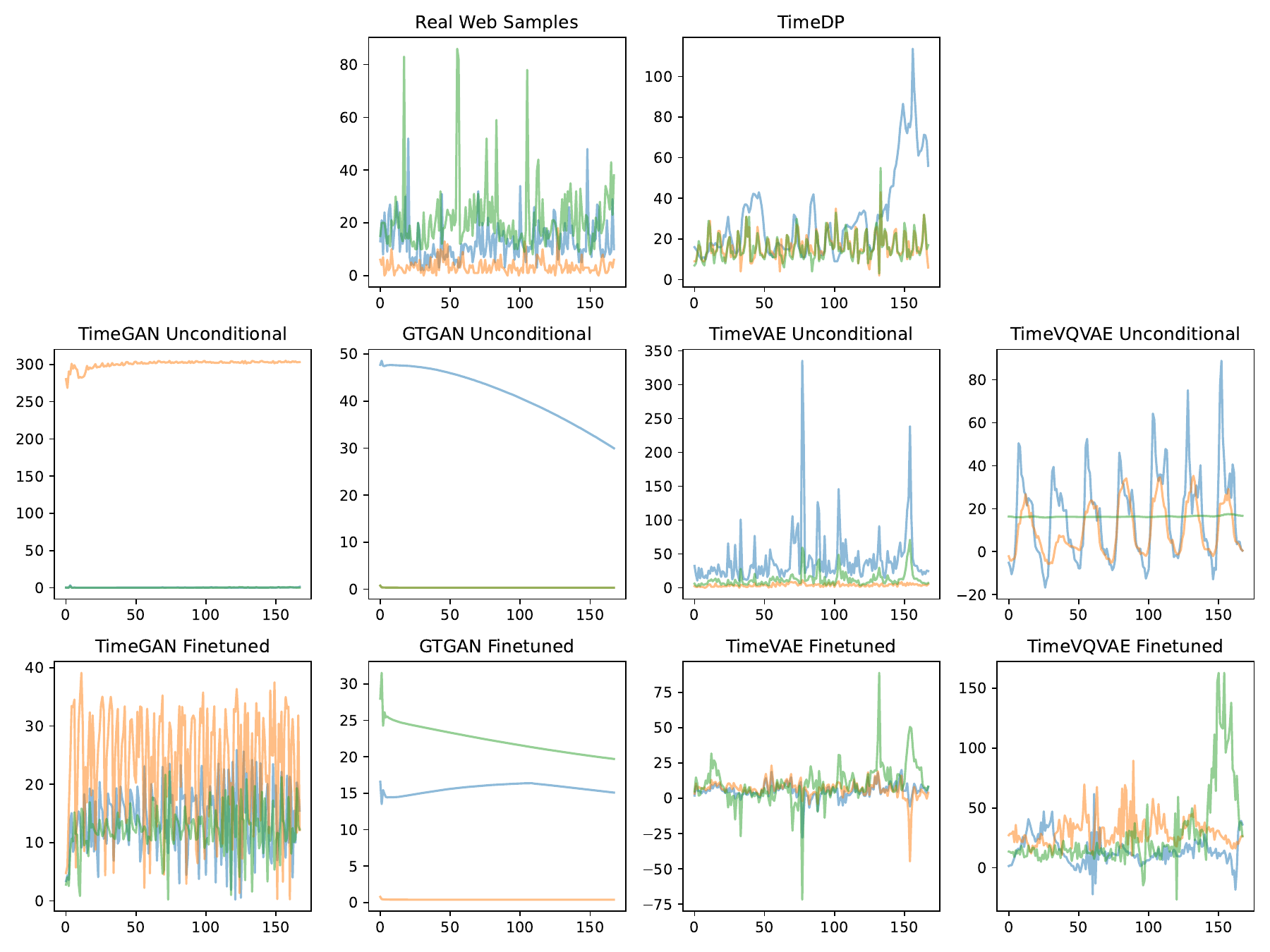}
  \caption{10-shot generation results of Web dataset, compared with real Web dataset samples.}
  \label{fig:web10_shots}
\end{figure*}

%% file: src_figtabs/fig_stock_zs.tex
\begin{figure*}[!t]
  \centering
  \includegraphics[width=0.95\textwidth]{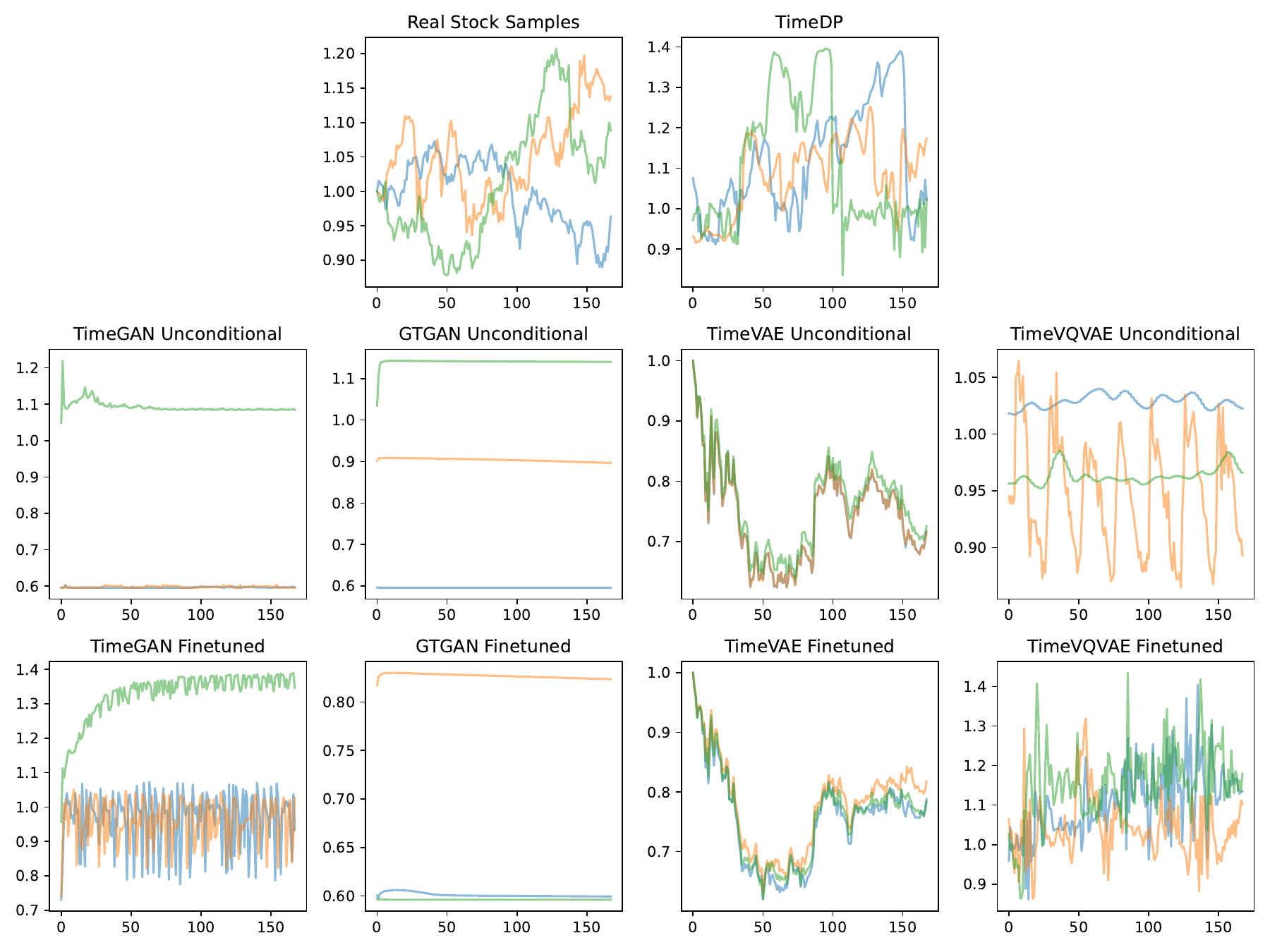}
  \caption{10-shot generation results of Stock dataset, compared with real Stock dataset samples.}
  \label{fig:stock10_shots}
\end{figure*}